\pdfoutput=1
\documentclass[11pt,a4paper]{article}
\usepackage[T1]{fontenc}

\usepackage{graphicx}
\graphicspath{{figures/}}
\usepackage{booktabs}
\usepackage{multirow}
\usepackage{tabularx}
\usepackage{placeins}
\usepackage{float}
\usepackage{enumitem}
\usepackage{amsmath,amssymb,mathtools}
\usepackage{times}
\usepackage{mathptmx}
\SetSymbolFont{letters}{normal}{OML}{txmi}{m}{it}
\SetSymbolFont{letters}{bold}{OML}{txmi}{bx}{it}
\usepackage{bm}
\usepackage[protrusion=true,expansion=true,tracking=false]{microtype}
\DeclareMathAlphabet{\cmcal}{OMS}{cmsy}{m}{n}
\newcommand{\loss}{\cmcal{L}}
\newcommand{\cropop}{\cmcal{C}}

\usepackage[preprint]{acl}
\usepackage{etoolbox}
\AtBeginEnvironment{equation}{\fontsize{10}{12}\selectfont}
\usepackage[all]{hypcap}
\usepackage[nameinlink,capitalize,noabbrev]{cleveref}

\usepackage{tikz}
\usetikzlibrary{arrows,calc,fit,positioning,shapes.geometric}

\setlist[itemize]{leftmargin=1.5em,topsep=4pt,itemsep=3pt,parsep=0pt}

\newcommand{\method}{EgoWholeView}

\newif\ifshowunfinishedexperiments
\showunfinishedexperimentsfalse % set true only after replacing all TBD cells

\definecolor{egoblue}{RGB}{37,94,158}
\definecolor{egoteal}{RGB}{40,143,130}
\definecolor{egogreen}{RGB}{82,150,84}
\definecolor{egored}{RGB}{181,83,72}
\definecolor{egogold}{RGB}{197,146,45}
\definecolor{egogray}{RGB}{86,91,100}
\hypersetup{linkcolor=darkblue,citecolor=darkblue,urlcolor=darkblue}

\tikzset{
  egobox/.style={draw=egogray!80, rounded corners=1.5pt, thick, align=center, inner xsep=6pt, inner ysep=5pt, font=\footnotesize, fill=white},
  egostage/.style={egobox, minimum width=1.35in, minimum height=0.36in},
  egoarrow/.style={->, >=latex, thick, draw=egogray!85},
  egodata/.style={egobox, fill=egoblue!8, draw=egoblue!75},
  egomodel/.style={egobox, fill=egoteal!8, draw=egoteal!75},
  egoout/.style={egobox, fill=egogreen!9, draw=egogreen!75},
  egoeval/.style={egobox, fill=egogold!12, draw=egogold!75},
  egowarn/.style={egobox, fill=egored!8, draw=egored!75},
  egopanel/.style={draw=egogray!70, rounded corners=2pt, thick, fill=egogray!4, minimum height=0.86in, align=center, font=\footnotesize},
  egosmall/.style={font=\normalsize, align=center}
}

\newcommand{\egocam}[4]{%
  \begin{scope}[shift={(#1,#2)}, rotate=#3, color=#4]
    \fill (-0.1,-0.06) rectangle (0.02,0.06);
    \fill (0.02,-0.03) -- (0.1,-0.075) -- (0.1,0.075) -- (0.02,0.03) -- cycle;
  \end{scope}}

\title{\method: Camera-Only Head-Mounted Capture of\\
Body and Bimanual Hand Motion for Embodied Demonstrations}

\author{%
  \textbf{Yaoqing Hu}\textsuperscript{1},\quad
  \textbf{Xuguo He}\textsuperscript{2},\quad
  \textbf{Daohan Cui}\textsuperscript{3},\quad
  \textbf{Yukun Shi}\textsuperscript{4},\quad
  \textbf{Kai Chen}\textsuperscript{2,4,\ensuremath{\dagger}}\\[8pt]
  {\small\normalfont\textsuperscript{1}Zhongguancun Academy\qquad
   \textsuperscript{2}DeepCybo}\\[2pt]
  {\small\normalfont\textsuperscript{3}Beijing University of Chemical Technology}\\[2pt]
  {\small\normalfont\textsuperscript{4}Zhongguancun Institute of Artificial Intelligence}%
}

\newcommand{\egoteaser}{%
\begin{figure*}[t]
  \centering
  \includegraphics[width=\linewidth]{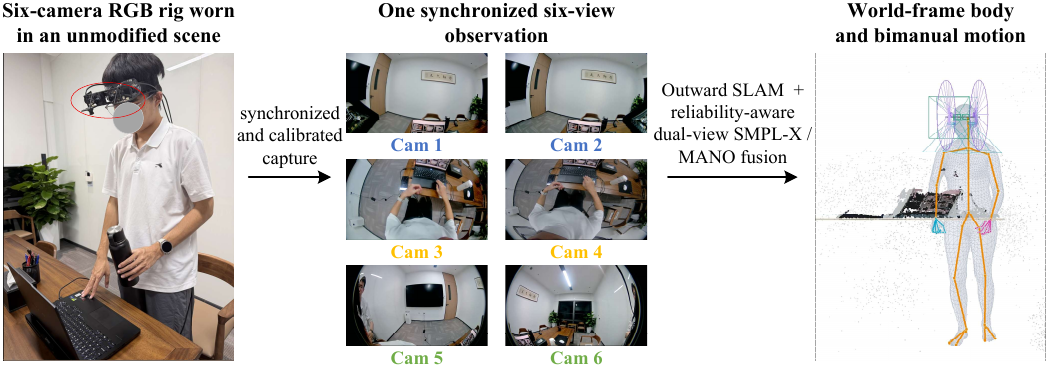}
  \caption{Overview of \method{}. Six synchronized cameras on a wearable rig provide observations for continuous outward-view multi-camera trajectory estimation and downward-view body and bimanual pose estimation. Paired downward-view predictions undergo reliability-aware fusion in the canonical space, followed by temporal refinement of local rotations. The system exports world-frame body and bimanual motion.}
  \label{fig:teaser}
\end{figure*}
}

\newcommand{\egopaperabstract}{%
\begingroup

Head-mounted body and bimanual motion capture supports self-avatars, teleoperation, and embodied demonstrations. However, a single egocentric camera is vulnerable to a limited field of view, scale ambiguity, and hand--object occlusion. We present \method{}, a wearable capture system of six synchronized wide-angle RGB cameras with a role-specialized pipeline. The outward views localize the rig in a world frame, while an overlapping downward pair independently predicts SMPL-X body and MANO hand parameters. Both predictors are adapted to egocentric fisheye data. To combine the two, reliability cues measured from the predictions themselves weight the fusion of body and hand joint rotations, while shape is aggregated across views and frames. The pair's baseline fixes metric scale, and triangulation recovers root translation. Relative to monocular prediction, fusion reduces PA-MPJPE by 3.7\% for the body and 11.2\% for the hands, and temporal refinement lowers angular-acceleration magnitude by 32.2\% and 31.0\%, respectively. On our six-camera recording, the pipeline produces body and hand states in a shared world frame using benchmark-fitted gates without retuning, and temporal refinement lowers hand acceleration magnitude by 41.08\%. Using only RGB cameras and no third-person view, \method{} offers a low-cost, portable route to egocentric interaction data.
\par
\endgroup
}

\hypersetup{
  pdftitle={EgoWholeView: Camera-Only Head-Mounted Capture of Body and Bimanual Hand Motion for Embodied Demonstrations},
  pdfauthor={Yaoqing Hu, Xuguo He, Daohan Cui, Yukun Shi, Kai Chen},
  pdfsubject={arXiv preprint},
  pdfkeywords={virtual reality, egocentric vision, wearable motion capture, hand tracking, embodied demonstrations}
}

\begin{document}

\maketitle
\begingroup
  \renewcommand{\thefootnote}{\fnsymbol{footnote}}%
  \footnotetext[2]{Corresponding author.}
\endgroup
\begin{abstract}
  \egopaperabstract
\end{abstract}
\egoteaser

\section{Introduction}

Body and bimanual hand motion are core inputs to virtual reality and 3D user interfaces~\cite{bowman2004dui}. Self-avatars and embodied telepresence~\cite{tome2019xregopose}, hand-driven interaction in VR~\cite{han2022umetrack}, and robot demonstration capture~\cite{khazatsky2024droid} all benefit from temporally coherent motion in a stable metric frame together with explicit reliability information~\cite{kolotouros2021prohmr,bramlage2023plausible}. Existing capture paradigms present a deployment--observability trade-off. External multi-camera systems are accurate but constrain the capture volume~\cite{ionescu2014human36m,iskakov2019learnable}. Monocular egocentric RGB enables mobile capture but remains geometrically underconstrained: a single view covers only part of the near body, leaves metric scale unobservable, and is frequently interrupted by hand--object occlusion~\cite{prakash2024wildhands,mucha2026vhpot}. Moving hands can also degrade SLAM, and out-of-view motion requires explicit infilling~\cite{zhang2025hawor}. Recovering both hands together with the arms and torso is harder than isolated hand regression, and from a single head-mounted camera the body is often outside the reliable field of view (FOV).

Most commercial egocentric headsets already carry several cameras, but their sensing layouts are not designed for complete body capture, and such platforms remain costly or research-restricted. For example, HOT3D uses Meta Quest 3 for hand--object tracking alone~\cite{banerjee2024hot3d}. Recovering body and hand motion in a world frame requires two measurements that a single forward-facing view cannot supply: the metric pose of the rig itself in the world frame, and the metric geometry of the body and hands relative to that rig. The two measurements rely on different observations: the first on static scene features, the second on moving body parts. Camera identity, coordinate frame, timestamp, and the body-or-scene distinction therefore belong with every intermediate prediction. Overlapping views of the same joint are rarely equally reliable, so both contribute rather than one being selected.

In this work, we meet these requirements with a portable head-mounted RGB rig alone, through the role-specialized architecture of \Cref{fig:teaser}. The rig carries six synchronized wide-angle cameras: forward and lateral outward views jointly recover its pose in a world frame, and an overlapping downward pair observes the near body and both hands. Released regressors target distant, weakly distorted views, which do not match the close-range, strongly distorted images that a head-mounted fisheye rig produces. We therefore adapt both predictors using egocentric recordings and synthetic fisheye images.

To combine the two downward views without assuming that either is reliable, the branches fuse the paired local rotations, with per-joint reliability weights deciding how much each view contributes. Fusion takes place in the canonical space of SMPL-X~\cite{pavlakos2019expressive} and MANO~\cite{romero2017embodied}, which is independent of any camera. Global orientation and monocular root translation are expressed in each camera's own frame and are handled separately: orientation is fused on $SO(3)$ once both views are brought into the frame of one downward camera, whereas metric root translation is recovered from calibrated stereo semantic-joint correspondences rather than from the two monocular root translation predictions. Articulation is evaluated under similarity-aligned protocols, and metric placement is demonstrated on the rig.

\noindent\textbf{Contributions.}
Our main contributions are summarized as follows:
\begin{itemize}
\renewcommand{\labelitemi}{\scriptsize$\bullet$}
\setlength{\itemsep}{2pt}
\setlength{\parskip}{0pt}
\item \textbf{Role-Specialized Six-Camera Wearable Rig:} a synchronized, calibrated head-mounted rig whose outward views recover its pose while an overlapping downward pair observes the near body and both hands. Body and left/right hand assignments are retained throughout.
\item \textbf{Egocentric Adaptation and Canonical-Space Fusion:} the SMPL-X and MANO predictors are adapted to egocentric fisheye views; their paired outputs are fused at the parameter level in the canonical space under per-joint reliability gates, and the fused rotations receive a disagreement-adaptive $SO(3)$ refinement to improve temporal coherence.
\item \textbf{Benchmark Evaluation and Real-Rig Demonstration:} the method lowers PA-MPJPE by 3.7\% on UnrealEgo-RW~\cite{akada2022unrealego} and 11.2\% on DexYCB~\cite{chao2021dexycb} relative to monocular prediction, applies the DexYCB-fitted hand gate to HO3Dv3~\cite{hampali2020honnotate} without refitting, and reduces angular-acceleration magnitude by 32.2\% for the body and 31.0\% for the hands. On our own six-camera recording, gates frozen from the benchmarks are used without retuning, and the pipeline runs end to end.
\end{itemize}

\section{Related Work}

\subsection{Wearable Egocentric Capture for Embodied Applications}

Embodied applications constrain both the content and the representation of captured motion. A self-avatar driven by the user's own movement can induce ownership of and agency over a virtual body~\cite{slater2010bodytransfer,kilteni2012embodiment}. This effect is not fixed by appearance: personalization raises ownership and presence~\cite{waltemate2018avatar}, tracking fidelity changes embodiment~\cite{eubanks2020trackingfidelity}, and motion congruence shapes self-recognition~\cite{kim2024selfavatar}.

Camera-based hand tracking is an established VR input~\cite{han2020megatrack,han2022umetrack} but degrades during physical-object interaction~\cite{xu2026surfacexr,cho2026unifiedho}, and how it fails matters as much as its mean error: jitter and dropouts propagate into avatar motion~\cite{maiorca2024artifacts}. Capture inherits this difficulty because a recording system cannot require the wearer to keep both hands visible. Demonstration capture additionally requires a stable world frame~\cite{zhang2018vrteleop}. Together, these applications call for temporally coherent, metric, parametric body-and-hand motion, with explicit reliability information~\cite{kolotouros2021prohmr,bramlage2023plausible}.

Existing wearables meet subsets of these needs. Project Aria is the closest platform, exposing calibrated outward fisheye streams with metric device tracking~\cite{engel2023projectaria} and anchoring work on hand--object benchmarking~\cite{banerjee2024hot3d} and dataset-scale full-body motion~\cite{ma2024nymeria}. These platforms observe the scene, so the wearer's body reaches the model indirectly. EgoAllo accepts that constraint and recovers allocentric body-and-hand motion from head-mounted sensing~\cite{yi2025egoallo}; EgoBody targets social interaction~\cite{zhang2022egobody}. Downward-facing rigs take the opposite position, maximizing direct wearer visibility~\cite{xu2019mo2cap2,tome2019xregopose,zhao2021egoglass,akada2022unrealego}. Either way the world frame comes from camera-only SLAM, now including generalized-camera tracking~\cite{campos2021orbslam3,urban2016multicolslam,demayo2025monado}.

Outward world sensing and downward body capture remain separate lines of work. To our knowledge, no wearable dedicates outward cameras to rig trajectory and an overlapping downward pair to directly observed body and hands, which is the system-level gap our six-camera design targets. Egocentric corpora already train embodied models and data engines~\cite{grauman2022ego4d,wang2023holoassist,nair2022r3m,brohan2023rt2,zhang2025robowheel}, but generally lack synchronized, metric, parametric body-and-hand states for motion supervision and retargeting.

\subsection{Parametric Body and Hand Reconstruction}

MANO and SMPL-X provide structured parametric representations for hands and expressive bodies, respectively: subject-specific shape and local joint rotations defined relative to kinematic trees~\cite{romero2017embodied,pavlakos2019expressive}. Shape can be personalized to a wearer, and local rotations transferred to an articulated avatar more directly than loose keypoints. This factorization is important for a moving rig: the same articulation may be predicted from two cameras even though its global orientation and translation differ between camera frames. Transformer regressors produce strong monocular hand hypotheses from a single crop~\cite{pavlakos2023hamer,potamias2025wilor,lin2026geohand}. Egocentric variants adapt these models to the close range, distortion, truncation, blur, and hand--object occlusion produced by head-mounted cameras~\cite{prakash2024wildhands,mucha2026vhpot}. Egocentric whole-body regressors extend the same adaptation to the body~\cite{na2026egowholebody}, where the domain gap is wider: downward views strongly foreshorten and crop the torso and rarely observe the legs.

On a moving wearer these regressors face two further difficulties. World-space reconstruction must separate camera and wearer motion because each prediction is expressed in a frame that is itself moving~\cite{zhang2025hawor}. Independently regressed frames also introduce jitter. Existing temporal methods learn body sequence priors or add video context for hand reconstruction~\cite{kocabas2020vibe,lin2024omnihands}, but couple temporal consistency to a sequence predictor and therefore change the deployed model. Keeping a released regressor frozen leaves only its per-frame output rotations to refine.

\subsection{Reliability-Aware Multi-View Fusion}

One camera may hold the wrist while losing the fingertips, and the other may fail on different joints. ProHMR and HuManiFlow model distributions over plausible bodies, POCO predicts per-sample uncertainty, and such estimates may still require post-hoc calibration~\cite{kolotouros2021prohmr,sengupta2023humaniflow,dwivedi2024poco,bramlage2023plausible}. PARE and HandOccNet attend to occluded parts~\cite{kocabas2021pare,park2022handoccnet}; learnable triangulation scores camera--joint pairs before geometric aggregation~\cite{iskakov2019learnable}. A single score for an entire view is therefore too coarse. The closest baselines fuse elsewhere: S2DHand converts two 21-joint predictions into pseudo-labels that update a single-view estimator~\cite{liu2024s2dhand}, whereas Rule Meets Learning combines learned confidence with a fixed rule over per-view poses~\cite{ren2025rule}. Neither assigns view weights directly to MANO local rotations, the quantity that must be combined per joint.

Other multi-view methods fuse features or intermediate geometry. UmeTrack sits closest to deployment, processing multiple views and frames jointly to predict one world-space hand pose~\cite{han2022umetrack}. Others embed cross-view evidence before regression: in image space, in a shared 3D point representation, or over a graph of joints and quality cues~\cite{tse2023spectral,yang2023poem,ren2022mining}. HaMuCo co-trains a single-view estimator through cross-view interaction~\cite{zheng2023hamuco}, and annotation pipelines fit one MANO state to synchronized image and depth evidence~\cite{hampali2020honnotate}.

All of them need a trainable multi-view network, cross-view features, or iterative optimization. We instead study parameter-level fusion over released monocular regressors that are adapted with single-view supervision only and then frozen for gate fitting, fusion, and evaluation, so their weights never see a multi-view objective. Our fusion stage neither updates the monocular backbones nor performs cross-view feature fusion. It operates on the predicted parameters, reading intermediate features within a single view only, to derive one reliability cue. What the two views share is their predicted SMPL-X or MANO parameters, so we fuse local rotations in the camera-independent canonical space defined in \Cref{sec:parametric_fusion}. Triangulation is used only for metric root placement, not articulation. Recent rigs and datasets with similar sensor layouts still lack an established paired-MANO protocol~\cite{rim2025egoexo,liu2022hoi4d,ropedia2026xperience10m}.

\section{Six-Camera Wearable Capture System}

Each camera is assigned a specific sensing role, and every prediction retains its coordinate frame. We fuse synchronized predictions only when their parameterizations are compatible. \Cref{tab:system_outputs} summarizes the frame, source cameras, and processing of each system output.

\begin{table*}[t]
\caption{System outputs, representations, coordinate frames, and processing.}
\label{tab:system_outputs}
\centering
\small
\begingroup
\setlength{\tabcolsep}{5pt}
\renewcommand{\tabularxcolumn}[1]{m{#1}}
\begin{tabularx}{\textwidth}{@{}>{\raggedright\arraybackslash}m{0.24\textwidth}>{\raggedright\arraybackslash}m{0.28\textwidth}>{\centering\arraybackslash}m{0.13\textwidth}>{\raggedright\arraybackslash}X@{}}
\hline
Output & Representation / frame & Cam. & Step \\
\hline
Rig pose $\boldsymbol T^w_{\mathrm{rig},t}$ & $SE(3)$, rig to world & 1/2, 5/6 & multi-view SLAM \\
\hline
Body $X_t^{B}$ & SMPL-X, root relative & 3/4 & gated fusion + temporal \\
\hline
Hand $X_t^{h}$ & MANO, root relative & 3/4 & gated fusion + temporal \\
\hline
Placement $\mathbf{P}_t$ & root $SE(3)$, Cam3 frame & 3/4 & $SO(3)$ orientation fusion + stereo triangulation \\
\hline
\end{tabularx}
\endgroup
\end{table*}

\subsection{Camera Layout and Hardware Implementation}

\begin{figure*}[t]
  \centering
  \resizebox{\linewidth}{!}{%
  \begin{minipage}{7in}
  \fontsize{9}{11}\selectfont
  \renewcommand{\normalsize}{\fontsize{9}{11}\selectfont}
  \begin{minipage}[t]{0.24\linewidth}
    \centering
    \begin{minipage}[c][1.72in][c]{\linewidth}
      \centering
      \begin{tikzpicture}[x=1cm,y=1cm,font=\normalsize]
        \node[inner sep=0pt] (rigphoto) at (0,0)
          {\includegraphics[width=1.58in,height=1.16in,keepaspectratio]{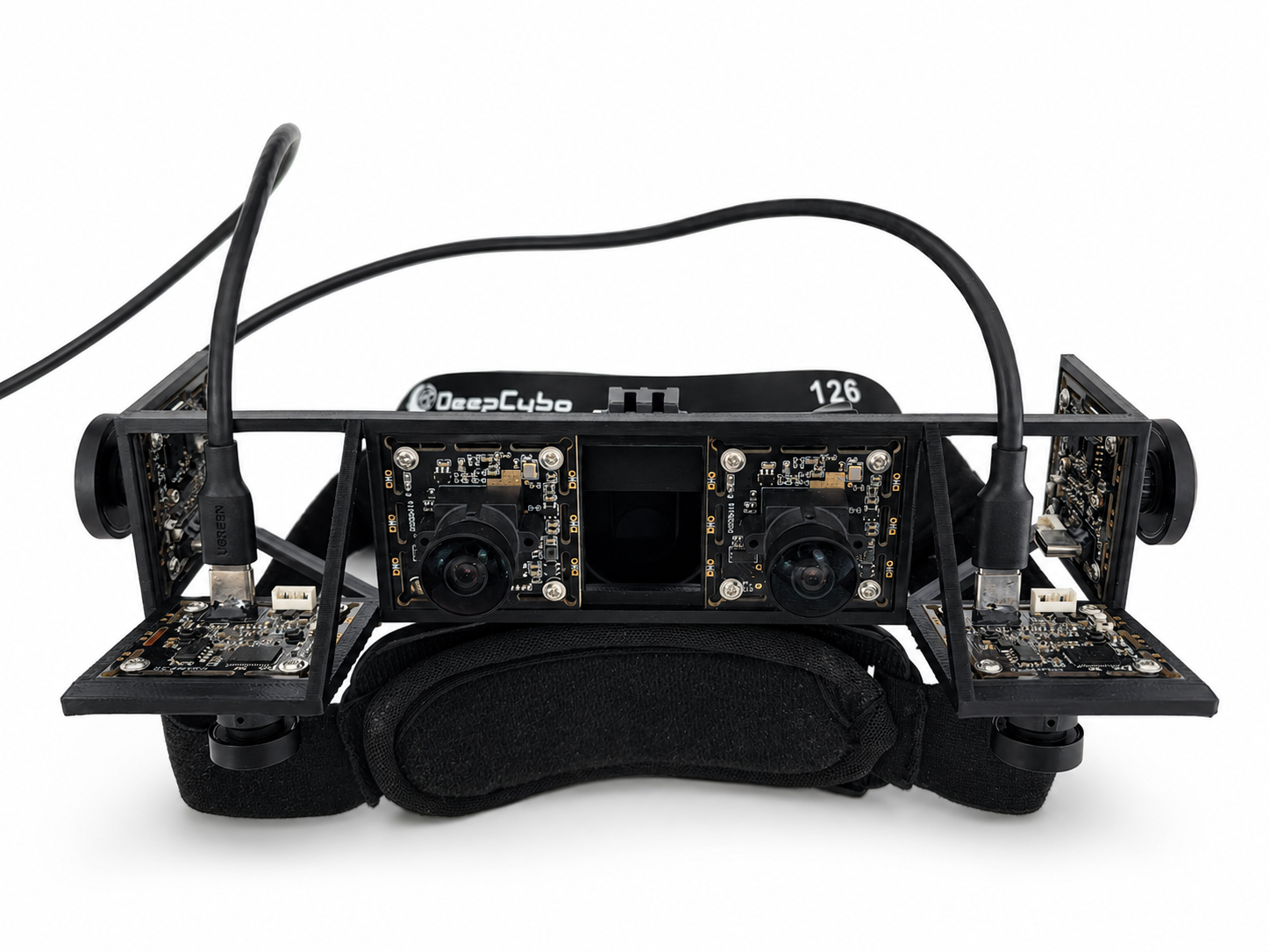}};
        \node[fill=white, fill opacity=0.92, text opacity=1, rounded corners=1pt, inner sep=1.5pt, text=egoblue] at (-1.28,1.12) {Cam1/2};
        \node[fill=white, fill opacity=0.92, text opacity=1, rounded corners=1pt, inner sep=1.5pt, text=egoteal] at (1.28,1.12) {Cam5/6};
        \node[fill=white, fill opacity=0.92, text opacity=1, rounded corners=1pt, inner sep=1.5pt, text=egogold!60!black] at (0,-1.12) {Cam3/4};
        \draw[->, >=latex, thick, draw=egoblue] (-1.05,0.92) -- (-0.52,0.28);
        \draw[->, >=latex, thick, draw=egoteal] (1.08,0.92) -- (1.45,0.18);
        \draw[->, >=latex, thick, draw=egogold!80] (0,-0.92) -- (0.28,-0.26);
      \end{tikzpicture}
    \end{minipage}
    \par\vspace{-8pt}
    {\normalsize (a)}
  \end{minipage}\hfill
  \begin{minipage}[t]{0.24\linewidth}
    \centering
    \begin{minipage}[c][1.72in][c]{\linewidth}
      \centering
      \begin{tikzpicture}[x=0.58cm,y=0.74cm]
      \fill[egoblue!14] (0.62,0.30) -- +(50:1.0) arc (50:-50:1.0) -- cycle;
      \draw[egoblue!55, dashed, thin] (0.62,0.30) -- +(50:1.0) (0.62,0.30) -- +(-50:1.0);
      \fill[egogold!22] (0.55,-0.15) -- +(-15:1.45) arc (-15:-115:1.45) -- cycle;
      \draw[egogold!70, dashed, thin] (0.55,-0.15) -- +(-15:1.45) (0.55,-0.15) -- +(-115:1.45);
      \fill[egogray!7] (0,0) circle (0.5);
      \draw[egogray, thick] (0,0) circle (0.5);
      \draw[egogray, thick] (0.48,0.06) -- (0.62,-0.02) -- (0.45,-0.12);
      \draw[egogray, line width=1.6pt] (14:0.58) arc (14:155:0.58);
      \draw[egogray, line width=1.2pt] (0.56,0.12) -- (0.52,-0.18);
      \draw[egogray, thick] (0,-0.5) -- (0,-0.72);
      \draw[egogray, thick] (0,-0.72) .. controls (0.2,-0.8) .. (0.35,-0.82);
      \draw[egogray, thick] (0,-0.72) .. controls (0.05,-1.3) .. (0.08,-1.85);
      \draw[egogray, thick] (0.35,-0.82) -- (0.62,-1.2) -- (1.0,-1.42);
      \draw[egogray, thick, fill=egogray!15] (1.06,-1.44) ellipse (0.14 and 0.09);
      \egocam{0.62}{0.30}{0}{egoblue}
      \egocam{0.55}{-0.15}{-65}{egogold}
      \node[egosmall, text=egoblue, anchor=west, align=left] at (0.80,1.20) {Cam1/2 frontal\\outward views};
      \node[egosmall, text=egogold!60!black, anchor=west, align=left] at (0.72,-0.66) {Cam3/4 downward pair\\(hands + near body)};
      \node[egosmall, text=egogray, anchor=west, align=left] at (0.64,-1.98) {hands at working\\distance};
      \end{tikzpicture}
    \end{minipage}
    \par\vspace{-8pt}
    {\normalsize (b)}
  \end{minipage}\hfill
  \begin{minipage}[t]{0.24\linewidth}
    \centering
    \begin{minipage}[c][1.72in][c]{\linewidth}
      \centering
      \begin{tikzpicture}[x=0.86cm,y=1.02cm]
      \fill[egoteal!14] (-0.6,0.08) -- +(130:0.85) arc (130:230:0.85) -- cycle;
      \fill[egoteal!14] (0.6,0.08) -- +(50:0.85) arc (50:-50:0.85) -- cycle;
      \draw[egoteal!55, dashed, thin] (-0.6,0.08) -- +(130:0.85) (-0.6,0.08) -- +(230:0.85);
      \draw[egoteal!55, dashed, thin] (0.6,0.08) -- +(50:0.85) (0.6,0.08) -- +(-50:0.85);
      \fill[egoblue!14, fill opacity=0.55] (-0.18,-0.18) -- +(-58:1.05) arc (-58:-122:1.05) -- cycle;
      \fill[egoblue!14, fill opacity=0.55] (0.18,-0.18) -- +(-58:1.05) arc (-58:-122:1.05) -- cycle;
      \draw[egoblue!55, dashed, thin] (-0.18,-0.18) -- +(-58:1.05) (-0.18,-0.18) -- +(-122:1.05);
      \draw[egoblue!55, dashed, thin] (0.18,-0.18) -- +(-58:1.05) (0.18,-0.18) -- +(-122:1.05);
      \fill[egogray!7] (0,0) circle (0.5);
      \draw[egogray, thick] (0,0) circle (0.5);
      \draw[egogray, line width=1.6pt] (6:0.58) arc (6:174:0.58);
      \draw[egogray, thick] (-0.9,-0.85) .. controls (-0.4,-0.62) and (0.4,-0.62) .. (0.9,-0.85);
      \foreach \fx in {-0.18,0.18}{
        \fill[egoblue!14] (\fx,-0.18) circle (0.085);
        \draw[egoblue, thick] (\fx,-0.18) circle (0.085);
        \fill[egoblue] (\fx,-0.18) circle (0.022);
      }
      \egocam{-0.6}{0.08}{180}{egoteal}
      \egocam{0.6}{0.08}{0}{egoteal}
      \egocam{-0.3}{-0.48}{-90}{egogold}
      \egocam{0.3}{-0.48}{-90}{egogold}
      \draw[stealth-stealth, egogold!70!black, thick] (-0.3,-0.76) -- (0.3,-0.76);
      \node[egosmall, text=egogold!60!black, anchor=west] at (0.48,-0.82) {baseline $b$};
      \node[egosmall, text=egoteal, anchor=east, align=right] at (-0.04,1.24) {Cam5 lateral\\(coverage)};
      \node[egosmall, text=egoteal, anchor=west, align=left] at (0.04,1.24) {Cam6 lateral\\(coverage)};
      \node[egosmall, text=egoblue] at (0,-1.48) {Cam1/2 frontal views};
      \end{tikzpicture}
    \end{minipage}
    \par\vspace{-8pt}
    {\normalsize (c)}
  \end{minipage}\hfill
  \begin{minipage}[t]{0.24\linewidth}
    \centering
    \begin{minipage}[c][1.72in][c]{\linewidth}
      \centering
      \begin{tikzpicture}[x=1cm,y=1cm,font=\normalsize]
        \node[inner sep=0pt, anchor=south west] at (0.00,0.02)
          {\includegraphics[height=1.50in]{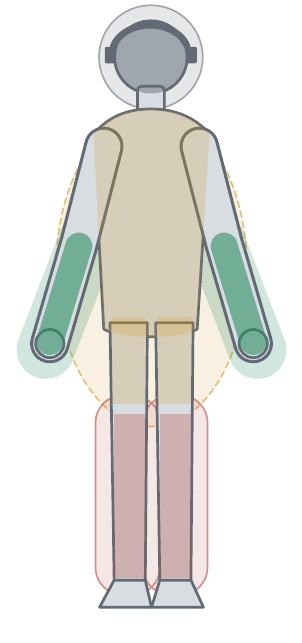}};
        \fill[egogreen] (1.78,3.08) rectangle (1.91,3.21);
        \node[anchor=west, align=left, text=egogreen!65!black] at (2.02,3.15)
          {paired hands\\+ forearms};
        \fill[egogold] (1.78,2.23) rectangle (1.91,2.36);
        \node[anchor=west, align=left, text=egogold!60!black] at (2.02,2.30)
          {variable\\near body};
        \fill[egored] (1.78,1.38) rectangle (1.91,1.51);
        \node[anchor=west, align=left, text=egored!75!black] at (2.02,1.45)
          {limited\\lower limbs};
        \fill[egogray] (1.78,0.53) rectangle (1.91,0.66);
        \node[anchor=west, align=left, text=egogray] at (2.02,0.60)
          {rig/self\\occlusion};
      \end{tikzpicture}
    \end{minipage}
    \par\vspace{-8pt}
    {\normalsize (d)}
  \end{minipage}
  \end{minipage}%
  }
  \caption{Six-camera wearable prototype and task-specialized layout. (a) Physical prototype. (b) Side view: the forward pair (Cam1/2) and the overlapping downward pair (Cam3/4). (c) Front view: the lateral pair (Cam5/6) and the downward-pair baseline $b$. (d) Downward-view evidence coverage; full-body prediction extends beyond the directly observed regions.}
  \label{fig:hardware_setup}
\end{figure*}

The rigid prototype carries three camera pairs, shown in \Cref{fig:hardware_setup}: a forward pair (Cam1/2), a lateral pair (Cam5/6), and an overlapping downward pair (Cam3/4). All six are commodity USB cameras with wide-angle optics spanning $170$--$220^\circ$ FOV. All four outward cameras supply the static-scene observations for trajectory estimation, with the lateral pair extending the coverage of the forward one. The downward pair observes the near body and both hands, supplying the paired hypotheses for reliability-aware fusion.

\subsection{System Overview and Capture Objective}

\begin{figure*}[t]
  \centering
  \includegraphics[width=\textwidth]{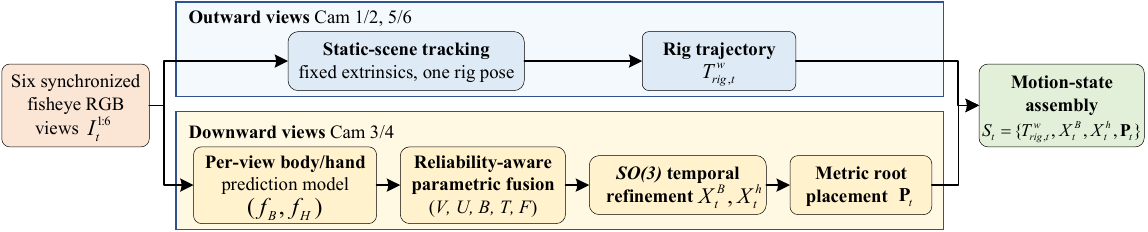}
  \caption{Processing overview. The outward branch recovers the rig trajectory; the downward branch predicts body and hand parameters per view. Paired predictions are fused in the canonical space: local rotations are weighted per joint, shape is aggregated across views and frames. The fused rotations are temporally refined and the roots placed metrically by triangulation.}
  \label{fig:pipeline}
\end{figure*}

\method{} integrates the role-specialized views into a portable capture pipeline, as shown in \Cref{fig:pipeline}. The rig trajectory establishes a common world reference, after which egocentric networks produce per-view body and hand hypotheses. Paired SMPL-X and same-side MANO predictions are fused in the canonical space using branch-specific reliability gates.

\subsection{Synchronization, Rig Calibration and Coordinate Frames}
\label{sec:calibration}

Each hardware trigger $t$ produces one timestamped frame per camera. Frames from the same trigger form a six-view observation, and their timestamps provide a synchronization consistency check. We calibrate the rigid assembly before capture using a cylindrical target adapted to the converging multicamera arrangement. \mbox{Its} curved marker layout follows self-identifying marker \mbox{designs}~\cite{wang2024cylindertag,hu2024lightweightnavigation}. This procedure estimates two quantities per view: a fisheye projection model $\pi_c$~\cite{scaramuzza2006flexible}, which maps a 3D point in camera $c$ coordinates to a pixel, and the extrinsic transform of that camera relative to the rig frame, which stays fixed because the assembly is rigid.

Cam1 defines the rig coordinate frame, and the downward pair uses Cam3 as its own reference for root placement. The fixed extrinsic transform
$\boldsymbol T^{\mathrm{rig}}_c$ maps points from camera $c$ to the rig frame, while
$\boldsymbol T^w_{\mathrm{rig},t}$ describes the time-varying rig pose in the world
frame. A point $\boldsymbol x^c$ expressed in camera $c$ coordinates is therefore
mapped to the world frame by
\begin{equation}
    \boldsymbol T^w_{c,t}=\boldsymbol T^w_{\mathrm{rig},t}\boldsymbol T^{\mathrm{rig}}_c,
    \;
    \boldsymbol x^w=\boldsymbol T^w_{c,t}\boldsymbol x^c .
    \label{eq:coordinate_transform}
\end{equation}
This transformation provides a common geometric reference for
camera-relative joints and vertices.

\subsection{Data Flow and System Outputs}
\label{sec:data_flow}
With the instance set $\mathbf I=\{B,L,R\}$, the exported state at time $t$ is:
\begin{equation}
\begin{aligned}
S_t&=\{\boldsymbol T^w_{\mathrm{rig},t},X_t^{B},\{X_t^{h}\}_{h\in\{L,R\}},\mathbf{P}_t\},\\
\mathbf{P}_t&=\{(\boldsymbol R_{\mathrm{root},t}^{i,o},\boldsymbol t_{\mathrm{root},t}^{i})\}_{i\in\mathbf I}.
\end{aligned}
\label{eq:state}
\end{equation}
Here $\boldsymbol T^w_{\mathrm{rig},t}$ is the rig pose in the world frame, $X_t^{B}$ contains temporally refined root-relative SMPL-X local rotations together with sequence-level shape, and $X_t^{h}$ contains temporally refined MANO local rotations together with sequence-level shape for hand side $h$. The placement term $\mathbf{P}_t$ contains the Cam3-frame root orientation and metric root translation of the body, left hand, and right hand, where $o$ marks fused quantities. If a hand is detected in only one downward view, its single-view MANO prediction is retained, but dual-view fusion and metric root placement are unavailable for that hand. Metric root placement is also unavailable when no valid triangulated joint remains. Per-view prediction, parameter fusion, temporal refinement, and root placement are described in \Cref{sec:egocentric_models,sec:parametric_fusion,sec:temporal_refinement,sec:metric_placement}.

\section{Method}
\label{sec:method}

\subsection{Rig Trajectory and World-Frame Recovery}
\label{sec:rig_trajectory}

Using the synchronized images and calibrated camera transforms, we estimate the 6-DoF rig trajectory $\boldsymbol T^w_{\mathrm{rig},t}$ from static-scene tracks after excluding the moving wearer. For a static landmark $\boldsymbol P_i^w$ observed at pixel $\boldsymbol u_{tci}$ in camera $c$, we define
\begin{equation}
\begin{aligned}
  e_{tci} &={}
  \left\|\pi_c\!\left(\boldsymbol T^c_{\mathrm{rig}}(\boldsymbol T^w_{\mathrm{rig},t})^{-1}\boldsymbol P_i^w\right)-\boldsymbol u_{tci}\right\|_2,\\
  \loss_{\mathrm{traj}} &={}
  \sum_{(t,c,i)\in O_{\mathrm{static}}}\rho\!\left(e_{tci}^{2}\right).
\end{aligned}
  \label{eq:rig_trajectory}
\end{equation}
where $\boldsymbol T^c_{\mathrm{rig}}=(\boldsymbol T^{\mathrm{rig}}_c)^{-1}$ is the rig-to-camera transform, $\pi_c$ is the camera projection model, $O_{\mathrm{static}}$ is the set of valid static observations, and $\rho$ is a robust penalty. Local bundle adjustment optimizes rig poses and landmarks jointly~\cite{urban2016multicolslam}, with loop closure supplied by a standard pose-graph backend~\cite{campos2021orbslam3}. Static tracks are aggregated across all four outward views; dynamic-region rejection removes the wearer where the arms and hands enter those images~\cite{zhang2025hawor}, since mapping the moving body would corrupt the trajectory. The first frame defines the world frame, so $\boldsymbol T^w_{\mathrm{rig},t}$ gives the rig pose relative to the recording origin.

\subsection{Egocentric Body and Hand Prediction Models}

\begin{figure*}[t]
  \centering
  \includegraphics[width=0.9110\textwidth]{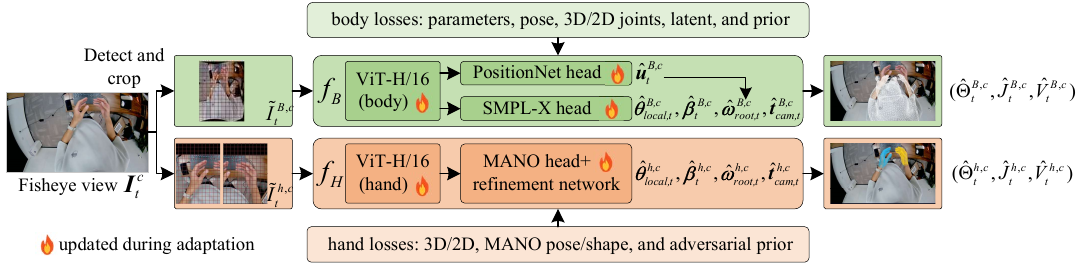}
  \vspace{-2pt}
  \caption{Per-view prediction and egocentric-domain adaptation. Body and hand crops from each downward view are processed by separate SMPL-X and MANO regressors, with branch-specific weights shared between Cam3 and Cam4. The loss boxes apply during adaptation only. At inference, paired Cam3/Cam4 hypotheses enter the independently parameterized body or hand fusion module in \Cref{sec:parametric_fusion}.}
  \label{fig:egocentric_network}
\end{figure*}
\label{sec:egocentric_models}

For each synchronized downward view $c\in\{3,4\}$, a body box $\boldsymbol b_t^{B,c}$ and a box $\boldsymbol b_t^{h,c}$ for each visible hand $h\in\{L,R\}$ define the two crops. Let $\cropop$ denote crop-and-resize preprocessing, with $\widetilde I_t^{B,c}=\cropop(I_t^c,\boldsymbol b_t^{B,c})$ and $\widetilde I_t^{h,c}=\cropop(I_t^c,\boldsymbol b_t^{h,c})$. The two branches process these crops independently before cross-view fusion, producing
\begin{equation}
\widehat\Theta_t^{B,c}=f_B(\widetilde I_t^{B,c}),\qquad \widehat\Theta_t^{h,c}=f_H(\widetilde I_t^{h,c}).
\label{eq:per_view_prediction}
\end{equation}

We instantiate the two branches with existing parametric regressors. Specifically, we use EgoSMPLX~\cite{na2026egowholebody} for SMPL-X body prediction and WiLoR~\cite{potamias2025wilor}, a ViT-based monocular hand regressor, for MANO prediction. Each branch predicts a parameter bundle $\widehat\Theta_t^{m,c}=\{\widehat{\boldsymbol\theta}_{\mathrm{local},t}^{m,c},\widehat{\boldsymbol\beta}_t^{m,c},\widehat{\boldsymbol\omega}_{\mathrm{root},t}^{m,c},\widehat{\boldsymbol t}_{\mathrm{cam},t}^{m,c}\}$, where $m\in\{B,H\}$ denotes the body or hand branch. Here, $\operatorname{Exp}(\boldsymbol\omega)$ maps an axis-angle vector $\boldsymbol\omega$ to a rotation matrix, while $\operatorname{Log}^{\vee}(\boldsymbol R)$ maps a rotation matrix back to its axis-angle vector. The SMPL-X or MANO layer first decodes root-relative joints and vertices,
$ (\widehat{\bar{\boldsymbol J}}_t^{m,c},\widehat{\bar{\boldsymbol V}}_t^{m,c})=M_m(\widehat{\boldsymbol\theta}_{\mathrm{local},t}^{m,c},\widehat{\boldsymbol\beta}_t^{m,c}) $. For every decoded point $n$, camera-frame placement is
$\widehat{\boldsymbol j}_{t,n}^{m,c}=\operatorname{Exp}(\widehat{\boldsymbol\omega}_{\mathrm{root},t}^{m,c})\widehat{\bar{\boldsymbol j}}_{t,n}^{m,c}+\widehat{\boldsymbol t}_{\mathrm{cam},t}^{m,c}$, and analogously for vertices. The body branch predicts 21 local SMPL-X rotations and the hand branch 15 local MANO rotations, each alongside one root rotation.

\noindent\textbf{Egocentric model adaptation.}
Both predictors retain their original architectures and are fine-tuned end to end on egocentric fisheye data. Both share the same three supervision families, namely validity-masked parameter losses on rotations and shape, an $\ell_1$ loss on decoded root-relative joints, and a reprojection loss on those joints. Neither supervises mesh vertices,
\begin{equation}
\loss_m=\lambda^m_{\mathrm{param}}\loss^m_{\mathrm{param}}+\lambda^m_{\mathrm{3D}}\loss^m_{\mathrm{3D}}+\lambda^m_{\mathrm{2D}}\loss^m_{\mathrm{2D}}+\loss^m_{\mathrm{prior}}.
\label{eq:egocentric_adaptation}
\end{equation}
They differ in two places. The body branch reprojects through the fisheye model of the sample's own camera and its crop transform, and adds an image-coordinate loss on the PositionNet head, whereas the hand branch follows WiLoR and reprojects in crop-normalized coordinates under a fixed virtual focal length, so fisheye distortion never enters its projection. The priors also differ: the hand branch uses WiLoR's adversarial pose--shape term, while the body branch adds a shape-norm prior, a DPoser-X diffusion prior whose weight decays from $0.1$ to $0.01$ over adaptation, and an explicit Smooth-$\ell_1$ loss on camera-frame root translation. Neither branch retunes the relative weights it inherits, and missing annotations are switched off by validity masks or zero weights. \Cref{fig:egocentric_network} shows the inference branches and adaptation-only supervision.

\subsection{Reliability-Aware Parametric Fusion}
\label{sec:parametric_fusion}
\begin{figure}[tbp]
\centering
\includegraphics[width=\linewidth]{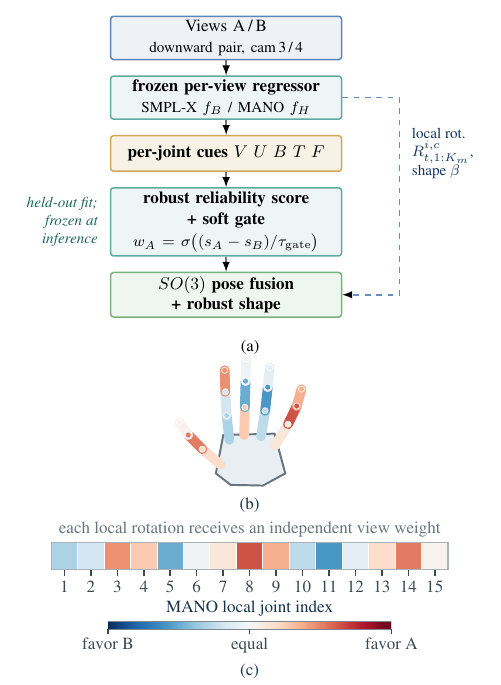}
\vspace{-3pt}
\caption{Reliability-aware parametric fusion. (a) Gate structure, shared by the body and hand branches but fitted with independent parameters; every normalized cue and aggregate score follows the higher-is-more-reliable convention. (b)--(c) Per-joint view preference and the resulting weights on a MANO hand.}
\label{fig:reliability_fusion}
\end{figure}
\Cref{fig:reliability_fusion} summarizes this stage. The two downward views provide paired hypotheses for one body instance and two hand instances. For $i\in\mathbf I$, the mapping $m(i)\in\{B,H\}$ selects its SMPL-X or MANO branch, with $K_H=15$ local MANO rotations and $K_B=21$ local SMPL-X body rotations in matrix form. A per-view hypothesis is
\begin{equation}
\widehat\Theta_t^{i,c}=\{\widehat{\boldsymbol R}_{t,1:K_{m(i)}}^{i,c},\widehat{\boldsymbol\beta}_t^{i,c},\widehat{\boldsymbol\omega}_{\mathrm{root},t}^{i,c},\widehat{\boldsymbol t}_{\mathrm{cam},t}^{i,c}\},\quad c\in\{3,4\}.
\label{eq:parametric_output}
\end{equation}
The first two terms describe local joint rotations and shape parameters. Both are defined relative to the kinematic tree rather than to any camera, and we refer to the space they span as the canonical space. The last two terms place the prediction relative to its source camera and are substantially less reliable under close-range truncation. This subsection therefore fuses local rotations and shape only. Root orientation and monocular root translation are handled in \Cref{sec:metric_placement}.

\noindent\textbf{Independent body and hand reliability gates.}
The body and hand branches use the same reliability-gate formulation but fit their parameters independently. Below, $c\in\{3,4\}$ indexes the downward camera, $j\in\{1,\ldots,K_m\}$ a local joint, and $\mathbf{K}=\{V,U,B,T,F\}$ the five cue types. For each indexed prediction, the frozen branch supplies
\begin{equation}
\boldsymbol{x}_{t,j}^{i,c}=\big[V_{t,j}^{i,c},U_{t,j}^{i,c},B_{t,j}^{i,c},T_{t,j}^{i,c},F_{t,j}^{i,c}\big]^\top,
\label{eq:reliability_cues}
\end{equation}
where $V$ is joint and kinematic-chain visibility, $U$ test-time perturbation dispersion, $B$ crop-boundary support, $T$ temporal inconsistency, and $F$ local feature dispersion. The symbol $x_{t,j,k}^{i,c}$ denotes the component associated with cue $k\in\mathbf{K}$. The $V$ and $B$ cues are computed from projected vertices affected by joint $j$, whereas $F$ is sampled at its projected descendant joints. The per-view root pose is used only for projection. The resulting scores set the view weights for paired local-rotation fusion.

The five raw cues differ in scale and reliability direction. We therefore map each cue to a normalized reliability value:
\begin{equation}
r_{t,j,k}^{i,c}=\sigma\!\left(d_k^m(x_{t,j,k}^{i,c}-\mu_k^m)/\eta_k^m\right),
\label{eq:reliability_calibration}
\end{equation}
where $\sigma(z)=(1+e^{-z})^{-1}$ is the logistic sigmoid; $d_k^m\in\{-1,+1\}$ orients each cue so that larger values always indicate higher reliability; and $\mu_k^m$ and $\eta_k^m>0$ are its robust center and scale in branch $m$. Hence, larger $r_{t,j,k}^{i,c}\in(0,1)$ indicates greater joint-view reliability.

Let $\boldsymbol\alpha^m=(\alpha_V^m,\alpha_U^m,\alpha_B^m,\alpha_T^m,\alpha_F^m)$ denote the fitted weights of the five cues for branch $m$. The joint-level reliability score and the two view weights are
\begin{equation}
\begin{aligned}
s_{t,j}^{i,c}&=\sum_{k\in\mathbf{K}}\alpha_k^m r_{t,j,k}^{i,c},
\\
\alpha_k^m&\geq 0,\quad \sum_{k\in\mathbf{K}}\alpha_k^m=1,\\
w_{t,j}^{i,3}&=\sigma\!\left((s_{t,j}^{i,3}-s_{t,j}^{i,4})/\tau_{\mathrm{gate}}^m\right),
\\
w_{t,j}^{i,4}&=1-w_{t,j}^{i,3}.
\end{aligned}
\label{eq:reliability_gate}
\end{equation}
Here $s_{t,j}^{i,c}\in(0,1)$ aggregates the five reliability values, while $\tau_{\mathrm{gate}}^m>0$ controls the sharpness of the view-weight assignment. With $\boldsymbol\mu^m=(\mu_k^m)_{k\in\mathbf{K}}$ and $\boldsymbol\eta^m=(\eta_k^m)_{k\in\mathbf{K}}$, the branch gate is $g_m=\{\boldsymbol\alpha^m,\tau_{\mathrm{gate}}^m,\boldsymbol\mu^m,\boldsymbol\eta^m\}$. The body gate $g_B$ and hand gate $g_H$ are fitted independently.

\noindent\textbf{Local-rotation and shape fusion.}
For $\boldsymbol R_A,\boldsymbol R_B\in SO(3)$, define
\[
\begin{aligned}
&\operatorname{GeoMean}_{SO(3)}(\boldsymbol R_A,\boldsymbol R_B;w)\\
&\qquad=\boldsymbol R_A\operatorname{Exp}\!\left(w\operatorname{Log}^{\vee}(\boldsymbol R_A^\top\boldsymbol R_B)\right).
\end{aligned}
\]
The fused local-pose and shape estimate is
\begin{equation}
\begin{aligned}
\boldsymbol R_{t,j}^{i,o}&=\operatorname{GeoMean}_{SO(3)}(\widehat{\boldsymbol R}_{t,j}^{i,3},\widehat{\boldsymbol R}_{t,j}^{i,4};w_{t,j}^{i,4}),\\
\boldsymbol\beta_t^i&=\underset{u\in N_t}{\operatorname{median}}\ \tfrac12(\widehat{\boldsymbol\beta}_u^{i,3}+\widehat{\boldsymbol\beta}_u^{i,4}),\\
(\bar{\boldsymbol J}_t^{i,o},\bar{\boldsymbol V}_t^{i,o})&=M_{m(i)}(\boldsymbol R_{t,1:K_{m(i)}}^{i,o},\boldsymbol\beta_t^i),
\end{aligned}
\label{eq:parametric_fusion}
\end{equation}
where the median is coordinate-wise and $N_t$ contains the frames of the same synchronized sequence. The superscript $o$ marks fused rotations and the quantities decoded from them, as opposed to the single-view predictions indexed by $c$; \Cref{sec:temporal_refinement} treats them as its observations. For each instance $i\in\mathbf I$, let $\boldsymbol\theta_{t,j}^{i,o}=\operatorname{Log}^{\vee}(\boldsymbol R_{t,j}^{i,o})$. The fused body and hand states are
\begin{equation}
\begin{aligned}
X_t^{B,o}&=\{\boldsymbol\theta_{t,1:K_B}^{B,o},\boldsymbol\beta_t^B,\bar{\boldsymbol J}_t^{B,o},\bar{\boldsymbol V}_t^{B,o}\},\\
X_t^{h,o}&=\{\boldsymbol\theta_{t,1:K_H}^{h,o},\boldsymbol\beta_t^h,\bar{\boldsymbol J}_t^{h,o},\bar{\boldsymbol V}_t^{h,o}\}.
\end{aligned}
\label{eq:mano_parameter_fusion}
\end{equation}
The two hand sides are processed independently.

\subsection{Adaptive \texorpdfstring{$SO(3)$}{SO(3)} Temporal Refinement}
\label{sec:temporal_refinement}
Frame-wise dual-view fusion can retain high-frequency rotation jitter. We temporally refine the $K_{m(i)}$ local rotations of each instance $i\in\mathbf I=\{B,L,R\}$. Sequence-level shape remains fixed, while root orientation and translation are not temporally filtered. For two joint-rotation matrices $\boldsymbol R_1,\boldsymbol R_2\in SO(3)$, their angular distance is $d(\boldsymbol R_1,\boldsymbol R_2)=\cos^{-1}((\operatorname{tr}(\boldsymbol R_1^\top\boldsymbol R_2)-1)/2)$. For joint $j$, cross-view disagreement, frame-to-frame motion, and fused reliability are $\Delta_{t,j}^i=d(\widehat{\boldsymbol R}_{t,j}^{i,3},\widehat{\boldsymbol R}_{t,j}^{i,4})$, $\nu_{t,j}^i=d(\boldsymbol R_{t-1,j}^{i,o},\boldsymbol R_{t,j}^{i,o})$, and $\bar{s}_{t,j}^i=\sum_{c\in\{3,4\}}w_{t,j}^{i,c}s_{t,j}^{i,c}$, respectively. These quantities set the two positive terms that a Kalman-style gain takes as process and observation variance:
\begin{equation}
\begin{aligned}
\sigma_{\mathrm{proc},t,j}^{2,i}
&=\sigma_{\mathrm{proc},0}^{2}
+\lambda_\nu\left(\nu_{t,j}^{i}/\sigma_\nu\right)^2,\\
\sigma_{\mathrm{obs},t,j}^{2,i}
&=\sigma_{\mathrm{obs},0}^{2}
+\lambda_{\mathrm{rel}}(1-\bar{s}_{t,j}^{i})^2
+\lambda_\Delta\left(\Delta_{t,j}^{i}/\sigma_\Delta\right)^2.
\end{aligned}
\label{eq:adaptive_temporal_variances}
\end{equation}
The subscript-$0$ terms are positive baselines, $\sigma_\nu,\sigma_\Delta>0$ are fixed angular scales shared by all adaptive candidates and never fitted, and the $\lambda$ terms are nonnegative contribution weights. These quantities proxy variance; they are not recursively propagated covariance estimates. Inspired by Kalman-style gains~\cite{kalman1960new} and speed-adaptive smoothing~\cite{casiez2012oneeuro}, we~compute a joint-wise current-observation weight and apply it through geodesic interpolation on $SO(3)$~\cite{park1997smooth}:
\begin{equation}
\begin{aligned}
\kappa_{t,j}^{i}
&=
\sigma_{\mathrm{proc},t,j}^{2,i}/
(\sigma_{\mathrm{proc},t,j}^{2,i}+\sigma_{\mathrm{obs},t,j}^{2,i}),\\
\boldsymbol R_{t,j}^{i,\mathrm{temp}}
&=\boldsymbol R_{t-1,j}^{i,\mathrm{temp}}
\operatorname{Exp}\left(
\kappa_{t,j}^{i}\operatorname{Log}^{\vee}\left(
(\boldsymbol R_{t-1,j}^{i,\mathrm{temp}})^\top\boldsymbol R_{t,j}^{i,o}
\right)\right).
\end{aligned}
\label{eq:so3_temporal_update}
\end{equation}
Fast motion raises $\sigma_{\mathrm{proc}}^2$ and thus $\kappa$, favoring the current observation and reducing lag. Low fused reliability or large cross-view disagreement raises $\sigma_{\mathrm{obs}}^2$ and lowers $\kappa$, favoring the previous refined state. In implementation, we bound the gain between $\kappa_{\min}=0.2$ and $\kappa_{\max}=0.98$ to avoid complete freezing or direct observation pass-through. The refined rotations replace the unfiltered local rotations before SMPL-X or MANO decoding; placement then uses the decoded articulation without temporally filtering root orientation or translation. We reset the state at sequence boundaries or non-consecutive frames and initialize the first valid frame as $\boldsymbol R_{t,j}^{i,\mathrm{temp}}=\boldsymbol R_{t,j}^{i,o}$. The recurrence is causal given its fused observations, but the full evaluated pipeline is offline because cue $T$ uses frame $t+1$ and shape is aggregated over the sequence.

\subsection{Global Orientation and Metric Root Placement}
\label{sec:metric_placement}
The two downward predictions express root orientation and monocular root translation in their respective camera frames. Because monocular root translation is unreliable under close-range truncation, we fuse orientation and recover metric root translation geometrically. Taking Cam3 as the reference, let $(\boldsymbol R_4^{3},\boldsymbol t_4^{3})$ denote the calibrated Cam4-to-Cam3 transform. The Cam3 orientation needs no transformation, whereas the Cam4 orientation is mapped by $\boldsymbol R_4^{3}$. The two are then averaged with equal weight,
\begingroup
\setlength{\abovedisplayskip}{3pt}
\setlength{\belowdisplayskip}{3pt}
\setlength{\abovedisplayshortskip}{2pt}
\setlength{\belowdisplayshortskip}{2pt}
\setlength{\jot}{0pt}
\begin{equation}
\begin{aligned}
\widetilde{\boldsymbol R}_{\mathrm{root},t}^{i,3}&=\operatorname{Exp}(\widehat{\boldsymbol\omega}_{\mathrm{root},t}^{i,3}),\\
\widetilde{\boldsymbol R}_{\mathrm{root},t}^{i,4}&=\boldsymbol R_4^{3}\operatorname{Exp}(\widehat{\boldsymbol\omega}_{\mathrm{root},t}^{i,4}),\\
\boldsymbol R_{\mathrm{root},t}^{i,o}&=\operatorname{GeoMean}_{SO(3)}\!\left(\widetilde{\boldsymbol R}_{\mathrm{root},t}^{i,3},\widetilde{\boldsymbol R}_{\mathrm{root},t}^{i,4};\tfrac12\right).
\end{aligned}
\label{eq:root_orientation_fusion}
\end{equation}
\endgroup
Translation is recovered from the two downward views rather than from monocular depth. Each semantic joint provides a pair of matching pixels in Cam3 and Cam4. Body pixels come from the PositionNet head, whereas hand pixels are decoded MANO joints mapped from crop coordinates back to the source images. We use all 21 MANO joints for each hand and the neck, shoulders, elbows, wrists, and hips for the body, omitting the weakly observed lower limbs. Using the calibrated fisheye camera models~\cite{scaramuzza2006flexible} and the known position and orientation between Cam3 and Cam4, we triangulate each pixel pair to obtain the 3D position $\boldsymbol X_{t,n}^{i}$ of joint $n$ in Cam3 coordinates. The known distance between the cameras fixes the absolute scale.

The decoded model provides the corresponding root-relative joint $\bar{\boldsymbol j}_{t,n}^{i}$. We discard triangulated joints with non-positive depth in either camera and let $N_{\mathrm{tri}}$ be the number of remaining joints. Once root orientation is fixed, joint $n$ estimates root translation as $\boldsymbol X_n-\boldsymbol R_{\mathrm{root}}\bar{\boldsymbol j}_n$. Omitting frame and instance indices for clarity, root translation is their arithmetic mean,
\begingroup
\setlength{\abovedisplayskip}{3pt}
\setlength{\belowdisplayskip}{3pt}
\setlength{\abovedisplayshortskip}{2pt}
\setlength{\belowdisplayshortskip}{2pt}
\begin{equation}
\boldsymbol t_{\mathrm{root}}=\frac{1}{N_{\mathrm{tri}}}\sum_{n=1}^{N_{\mathrm{tri}}}\left(\boldsymbol X_n-\boldsymbol R_{\mathrm{root}}\bar{\boldsymbol j}_n\right).
\label{eq:translation_alignment}
\end{equation}
\endgroup
If $N_{\mathrm{tri}}=0$, no root translation is produced. This stage otherwise updates root translation while leaving articulation, orientation, and shape fixed. Because the body regressor's own hand estimate is unreliable at this range, the fused MANO pose replaces the SMPL-X hand pose, with each hand orientation converted to the wrist-local frame. Each hand is then placed at its triangulated root, not at the wrist implied by the arm chain. The resulting poses are exported as $\mathbf{P}_t=\{(\boldsymbol R_{\mathrm{root},t}^{i,o},\boldsymbol t_{\mathrm{root},t}^{i})\}_{i\in\mathbf I}$ and transformed to the world frame by $\boldsymbol T^{w}_{i,t}=\boldsymbol T^{w}_{\mathrm{rig},t}\boldsymbol T^{\mathrm{rig}}_{3}\boldsymbol T^{3}_{i,t}$.

\section{Experiments and Discussion}
\label{sec:experiments}
% Structural headings use the single-column preprint style.

We evaluate body and hand fusion, adaptive $SO(3)$ temporal refinement, and the wearable's continuous trajectory and motion-capture outputs. The body and hand predictors are adapted independently, and their branches use separate gate parameters.

\subsection{Experimental Setup}
\label{sec:adaptation_settings}

The body and hand gates of \Cref{sec:parametric_fusion}, fitted independently, are called VUBTF after their five cues $V$, $U$, $B$, $T$, and $F$. Equal-weight fusion fixes $w_{t,j}^{i,c}=1/2$ in \Cref{eq:parametric_fusion}; VUBTF soft fusion uses the fitted gate weights. Both predictors, both gates, all cue-normalization statistics, and all temporal settings are frozen before testing.

\noindent\textbf{Body branch.}
The body predictor is fine-tuned from the released EgoSMPLX checkpoint, which uses the SMPLer-X-Huge backbone, on EgoPW~\cite{wang2022egopw}, SceneEgo~\cite{wang2023sceneego}, EgoWholeBody~\cite{na2026egowholebody}, UnrealEgo-RW~\cite{akada2022unrealego}, and synthetic EgoFishBody data. The combined adaptation set contains 551,924 valid samples per epoch, and fine-tuning runs for 12 epochs. All subsequent body experiments use the final checkpoint.

We fit the body gate's five global cue weights and one temperature for eight epochs on 2,048 UnrealEgo-RW stereo pairs sampled from 12 subjects using Adam (learning rate 0.05, batch size 128). After each epoch, the gate is evaluated on 512 pairs from S16, which are excluded from gate training, and the epoch with the lowest validation PA-MPJPE is retained, here the last of the eight. The selected gate is then frozen for UnrealEgo-RW test evaluation.

\noindent\textbf{Hand branch.}
The hand predictor is fine-tuned from the released WiLoR checkpoint on 391,003 cleaned HOI4D~\cite{liu2022hoi4d} samples and 155,000 synthetic EgoFishHands samples. Fine-tuning is performed using AdamW for 60 epochs (learning rate $10^{-5}$, weight decay $10^{-4}$, batch size 480), with the released loss weights $\lambda_{\mathrm{3D}}=0.05$, $\lambda_{\mathrm{2D}}=0.01$, $\lambda_{\mathrm{orient}}=\lambda_{\mathrm{pose}}=0.001$, and $\lambda_{\mathrm{shape}}=\lambda_{\mathrm{adv}}=0.0005$.

The hand gate fits the same six parameters: five global cue weights and one temperature. It is optimized for eight epochs on DexYCB S0 validation using Adam (learning rate 0.05, batch size 2,048). This fitting split contains 2,841 synchronized timestamps with three to eight valid views, yielding 21,497 unique images and 71,510 unordered pair appearances. The hand branch uses no held-out split for epoch selection, so the final-epoch gate and cue-normalization statistics are frozen for DexYCB testing and HO3Dv3 transfer.

\noindent\textbf{Inputs.}
On the benchmarks the body branch takes the full image and the hand branch uses the dataset-provided boxes, so no detector is involved in training or benchmark evaluation.

\noindent\textbf{Evaluation metric and gate parameters.}
For predicted and reference point matrices $\widehat{\boldsymbol X},\boldsymbol X^{*}\in\mathbb R^{N\times3}$, with one 3D point per row, define $r_n(s,\boldsymbol Q,\boldsymbol b)=\|s\widehat{\boldsymbol X}_n\boldsymbol Q^\top+\boldsymbol b-\boldsymbol X_n^{*}\|_2$. The standard similarity Procrustes alignment and the reported mean error are
\begin{equation}
\begin{aligned}
(s^{*},\boldsymbol Q^{*},\boldsymbol b^{*})
&=\underset{\substack{s>0,\,\boldsymbol Q\in SO(3)\\ \boldsymbol b\in\mathbb R^3}}{\arg\min}
\sum_{n=1}^{N}r_n(s,\boldsymbol Q,\boldsymbol b)^2,\\[-1pt]
E_{\mathrm{PA}}(\widehat{\boldsymbol X},\boldsymbol X^{*})
&=\frac{1}{N}\sum_{n=1}^{N}r_n(s^{*},\boldsymbol Q^{*},\boldsymbol b^{*}).
\end{aligned}
\label{eq:pa_mpjpe}
\end{equation}
PA-MPJPE applies $E_{\mathrm{PA}}$ to the joint matrices with $N=K$, and PA-MPVPE to the vertex matrices with $N=M$. For MANO, $(K,M)=(21,778)$, and for SMPL-X on UnrealEgo-RW, $K=15$. All distances are in millimeters.

\begin{table}[tbp]
\centering
\caption{Learned normalized cue weights, raw-cue reliability directions, and gate temperatures for the body and hand gates.}
\label{tab:gate_weights}
\small
\setlength{\tabcolsep}{3pt}
\begin{tabularx}{\columnwidth}{@{}Xlrr@{}}
\toprule
Cue & Reliable when & Body & Hand\\
\midrule
Visibility $V$ & Higher & \textbf{0.3228} & \textbf{0.3716}\\
Perturbation dispersion $U$ & Lower & 0.0912 & \textbf{0.3243}\\
Crop-boundary support $B$ & Higher & 0.1109 & 0.0029\\
Temporal inconsistency $T$ & Lower & \textbf{0.4551} & \textbf{0.2950}\\
Feature dispersion $F$ & Lower & 0.0201 & 0.0062\\
Gate temperature $\tau_{\mathrm{gate}}$ & --- & 0.1245 & 0.2285\\
\bottomrule
\end{tabularx}
\end{table}

\Cref{tab:gate_weights} lists the fitted coefficients. Although the two gates share one structure, they learn different cue combinations: the body gate emphasizes temporal consistency and visibility, whereas the hand gate emphasizes visibility, perturbation stability, and temporal consistency. The body's larger crop-boundary weight is consistent with torso and upper-arm truncation at the image border, whereas the annotated hand boxes are generally centered on the hand.

\subsection{Body Local-Pose Fusion on UnrealEgo-RW}

\begin{figure*}[t]
\centering
\includegraphics[width=\linewidth]{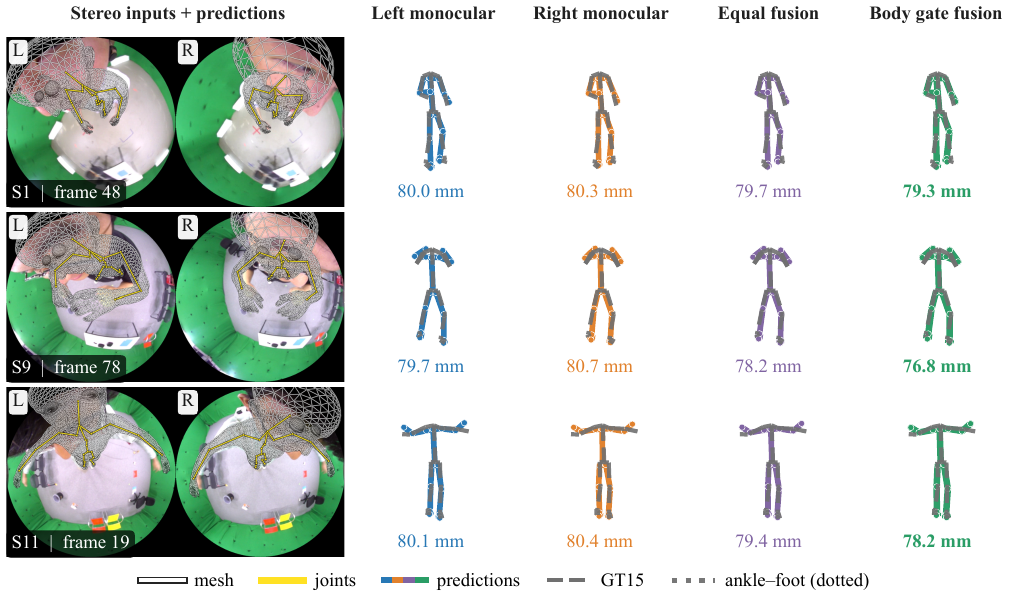}
\caption{Qualitative body local-pose fusion on UnrealEgo-RW. From left to right: stereo inputs with projected monocular SMPL-X predictions, left- and right-view predictions, equal-weight fusion, and VUBTF soft fusion. Predictions are independently Procrustes-aligned to GT15 (gray dashed), and values report PA-MPJPE in mm ($\downarrow$). Yellow joints and white wireframes show the projected predictions; dotted ankle--foot links indicate the heel--toe correspondences used for GT15 evaluation. The examples have monocular errors close to the test-set mean.}
\label{fig:body_fusion_qualitative}
\end{figure*}
\label{sec:body_evaluation}

\noindent\textbf{Data and protocol.}
The body evaluation uses the released UnrealEgo-RW test split: 93 synchronized stereo sequences from S1, S9, and S11, totaling 24,012 frame pairs (48,024 images), all held out from gate fitting, temporal-parameter fitting, and epoch selection.

Body joint accuracy uses GT15: the neck and bilateral shoulders, elbows, wrists, hips, knees, ankles, and ToeBase joints. Because SMPL-X lacks ToeBase joints, each annotation is approximately matched to the corresponding heel. Gate fitting excludes these two approximate matches and uses the remaining 13 joints. All reported test PA-MPJPE values use GT15. UnrealEgo-RW provides only sparse joint annotations, so we report joint error alone. The monocular baseline averages independently aligned per-view errors. Leg accuracy drops outside the downward views.

\begin{table*}[t]
\centering
\caption{SMPL-X fusion on the UnrealEgo-RW test split. PA-MPJPE uses GT15 and is given in mm.}
\label{tab:body_fusion_summary}
\begingroup
\small
\setlength{\tabcolsep}{5pt}
\begin{tabularx}{\textwidth}{@{}Xrrrrr@{}}
\toprule
Method & Mean & Median & P90 & P95 & \shortstack{Gain vs.\\mono. avg.}\\
\midrule
Left-view prediction & 79.6188 & 74.1331 & 102.4118 & 115.0452 & ---\\
Right-view prediction & 80.1623 & 74.5614 & 101.6197 & 116.6663 & ---\\
Monocular average & 79.8905 & 75.1588 & 101.2589 & 112.1287 & 0.0000\\
Equal-weight fusion & 77.3151 & 72.9845 & 96.8666 & 106.8710 & +2.5754\\
\textbf{VUBTF soft fusion} & \textbf{76.9409} & \textbf{72.6824} & \textbf{96.0671} & \textbf{106.3893} & \textbf{+2.9496}\\
\bottomrule
\end{tabularx}
\endgroup
\end{table*}

\noindent\textbf{Quantitative results.} In \Cref{tab:body_fusion_summary}, equal-weight fusion improves PA-MPJPE over the monocular average by 2.5754~mm and VUBTF soft fusion by 2.9496~mm (3.69\%), a further 0.3742~mm over equal-weight fusion. \Cref{fig:body_fusion_qualitative} shows qualitative examples.

\subsection{Reliability-Aware Dual-View MANO Fusion}
\label{sec:mano_fusion_results}

\noindent\textbf{Main DexYCB results.}
DexYCB S0 test contains 160 capture sequences holding 9,846 complete synchronized eight-view timestamps, or 78,768 images. All 28 camera combinations yield 275,688 pair appearances. Roles A and B are determined by camera serial number and do not denote fixed physical left and right positions. 

\begin{table*}[t]
\centering
\caption{Dual-view MANO fusion on DexYCB S0 test and frozen transfer to HO3Dv3 (mm). PA-MPJPE uses 21 joints and PA-MPVPE all 778 MANO vertices.}
\label{tab:hand_fusion}
\begingroup
\small
\setlength{\tabcolsep}{5pt}
\begin{tabularx}{\textwidth}{@{}Xrrrrr@{}}
\toprule
Method & Mean & Median & P90 & P95 & \shortstack{Gain vs.\\mono. avg.}\\
\midrule
\multicolumn{6}{@{}l}{\textit{PA-MPJPE (21 joints)}}\\[-1pt]
Mono A & 4.9092 & 4.3793 & 7.7093 & 9.1141 & ---\\
Mono B & 4.7478 & 4.2562 & 7.4034 & 8.8242 & ---\\
Monocular average & 4.8285 & 4.4214 & 7.2355 & 8.4284 & 0.0000\\
Equal-weight fusion & 4.3642 & 3.9486 & 6.7328 & 7.9130 & +0.4643\\
\textbf{VUBTF soft fusion} & \textbf{4.2872} & \textbf{3.8732} & \textbf{6.6027} & \textbf{7.7682} & \textbf{+0.5413}\\
\midrule
\multicolumn{6}{@{}l}{\textit{PA-MPVPE (778 vertices)}}\\[-1pt]
Mono A & 4.7557 & 4.2541 & 7.3973 & 8.7742 & ---\\
Mono B & 4.6186 & 4.1472 & 7.1306 & 8.4739 & ---\\
Monocular average & 4.6872 & 4.3023 & 6.9567 & 8.1240 & 0.0000\\
Equal-weight fusion & 4.2422 & 3.8499 & 6.4741 & 7.5945 & +0.4450\\
\textbf{VUBTF soft fusion} & \textbf{4.1741} & \textbf{3.7837} & \textbf{6.3589} & \textbf{7.4551} & \textbf{+0.5131}\\
\midrule
\multicolumn{6}{@{}l}{\textit{HO3Dv3 PA-MPJPE (21 joints, frozen transfer)}}\\[-1pt]
Mono A & 7.0625 & 6.8700 & 9.8260 & 10.8385 & ---\\
Mono B & 7.1124 & 6.7238 & 10.0545 & 11.1588 & ---\\
Monocular average & 7.0874 & 6.9924 & 9.4161 & 10.1860 & 0.0000\\
Equal-weight fusion & 6.4953 & 6.3483 & \textbf{8.7991} & \textbf{9.6039} & +0.5921\\
\textbf{VUBTF soft fusion} & \textbf{6.4775} & \textbf{6.3218} & 8.8179 & 9.6208 & \textbf{+0.6099}\\
\midrule
\multicolumn{6}{@{}l}{\textit{HO3Dv3 PA-MPVPE (778 vertices, frozen transfer)}}\\[-1pt]
Mono A & 6.9523 & 6.7141 & 9.6297 & 10.4536 & ---\\
Mono B & 7.0181 & 6.6324 & 9.8534 & 10.8022 & ---\\
Monocular average & 6.9852 & 6.8803 & 9.1858 & 9.9498 & 0.0000\\
Equal-weight fusion & 6.4274 & 6.3011 & \textbf{8.6324} & \textbf{9.3206} & +0.5578\\
\textbf{VUBTF soft fusion} & \textbf{6.4141} & \textbf{6.2617} & 8.6532 & 9.3921 & \textbf{+0.5711}\\
\bottomrule
\end{tabularx}
\endgroup
\end{table*}

\begin{figure}[tbp]
\centering
\includegraphics[width=\linewidth]{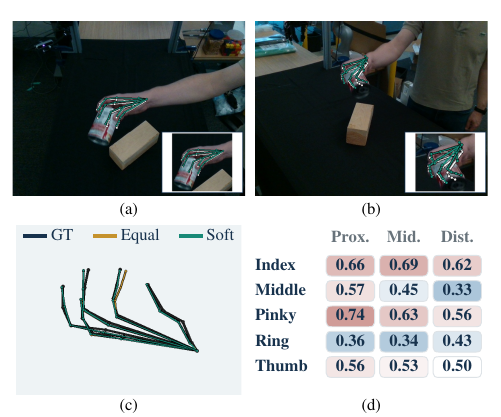}
\caption{Qualitative dual-view MANO fusion on DexYCB. (a)--(b) Synchronized views with monocular predictions and enlarged hand crops. (c) Ground-truth joints (GT), equal-weight fusion, and VUBTF soft fusion. (d) Per-joint view weights for proximal (Prox.), middle (Mid.), and distal (Dist.) finger joints.}
\label{fig:mano_fusion_qualitative}
\end{figure}

The official DexYCB MANO annotations are decoded into the reference meshes used for PA-MPVPE. As \Cref{tab:hand_fusion} shows, equal-weight fusion reduces PA-MPJPE and PA-MPVPE relative to the monocular average by 0.4643 and 0.4450~mm, respectively. With the learned reliability gate, the two errors decrease further to 4.2872 and 4.1741~mm, corresponding to reductions of 0.0770 and 0.0681~mm over equal-weight fusion. Across all 28 camera pairs, mean PA-MPJPE decreases by 0.0196--0.1645~mm. VUBTF soft fusion outperforms equal-weight fusion on 56.0\% of pair appearances and 139 of the 160 test sequences, with a sequence-block-bootstrap 95\% interval of $[0.0648,0.0895]$~mm. \Cref{fig:mano_fusion_qualitative} visualizes the predictions and per-joint weights.

\noindent\textbf{Frozen transfer to HO3Dv3.}
With the adapted WiLoR checkpoint and DexYCB-fitted gate frozen, we evaluate 4,854 synchronized HO3Dv3 frames spanning 19,242 of the 20,137 images in the official evaluation split, which yield 33,456 unordered pair appearances. The remaining 895 images have no synchronized multi-view counterpart. As on DexYCB, all metrics are computed locally from the released MANO annotations, which we decode into the 778-vertex reference meshes used for PA-MPVPE.

As \Cref{tab:hand_fusion} shows, equal-weight fusion improves the HO3Dv3 monocular average by 0.5921~mm (8.35\%) in PA-MPJPE and 0.5578~mm (7.99\%) in PA-MPVPE. VUBTF soft fusion achieves the lowest mean and median on both metrics, improving it by 0.6099~mm (8.61\%) and 0.5711~mm (8.18\%).

\subsection{Adaptive \texorpdfstring{$SO(3)$}{SO(3)} Temporal Refinement}
\label{sec:temporal_results}

The temporal module refines only local rotations, preserving sequence-level shape. With $\boldsymbol D_{t,j}=\boldsymbol R_{t-1,j}^\top\boldsymbol R_{t,j}$ and $\boldsymbol A_{t,j}=\boldsymbol D_{t,j}^\top\boldsymbol D_{t+1,j}$, the angular-velocity and angular-acceleration errors are $d(\widehat{\boldsymbol D},\boldsymbol D^{*})$ and $d(\widehat{\boldsymbol A},\boldsymbol A^{*})$, while acceleration magnitude is $d(\boldsymbol I,\widehat{\boldsymbol A})$ and needs no reference. All three average over local rotations and over transitions within a sequence, in $^\circ$/frame and $^\circ$/frame$^2$. UnrealEgo-RW has no ground-truth SMPL-X local rotations, so the body branch reports magnitude only.

\noindent\textbf{Hand branch.}
We select the temporal-refinement settings on DexYCB S0 validation and keep them fixed during testing. On 71,510 validation pair appearances, we compare six fixed gains (\(\kappa=0.50,\allowbreak 0.65,\allowbreak 0.75,\allowbreak 0.85,\allowbreak 0.90,\allowbreak 0.95\)) with three adaptive variants that compute \(\kappa\) for each frame and MANO joint. The variants differ in their baseline process and observation variances and in the weight on cross-view disagreement. The unrefined baseline obtains 4.0144 mm PA-MPJPE. The best fixed gain (\(\kappa=0.75\)) reduces it to 4.0013 mm, while the best adaptive variant reaches 3.9922 mm. The selected adaptive settings are
\[
\begin{aligned}
&(\sigma_{\mathrm{proc},0}^{2},\lambda_\nu,\sigma_{\mathrm{obs},0}^{2},\lambda_{\mathrm{rel}},\lambda_\Delta)\\
&\qquad=(1.0,0.5,0.1,0.25,0.5)
\end{aligned}
\]
and \((\sigma_\nu,\sigma_\Delta)=(0.12,0.25)\) rad. Both are frozen before testing.

\begin{table*}[t]
\centering
\caption{Temporal-refinement results on DexYCB S0 test. PA-MPJPE is in mm; angular metrics use $^\circ$/frame for velocity and $^\circ$/frame$^2$ for acceleration.}
\label{tab:temporal_test}
\begingroup
\small
\setlength{\tabcolsep}{5pt}
\begin{tabularx}{\textwidth}{@{}Xrrrr@{}}
\toprule
Method & \shortstack{PA-\\MPJPE} & \shortstack{Angular-\\velocity\\error} & \shortstack{Angular-\\acceleration\\error} & \shortstack{Acceleration\\magnitude}\\
\midrule
VUBTF observation & 4.2872 & 6.1053 & 8.1901 & 2.2269\\
Fixed gain, $\kappa=0.75$ & 4.2687 & 5.9673 & 7.8929 & 1.4968\\
\textbf{Adaptive gain} & \textbf{4.2621} & \textbf{5.9601} & \textbf{7.8861} & 1.5356\\
\bottomrule
\end{tabularx}
\endgroup
\end{table*}

On 275,688 test pair appearances (\Cref{tab:temporal_test}), adaptive refinement lowers the PA-MPJPE of the VUBTF observation, the fused estimate before refinement, by 0.0251~mm, with a sequence-block-bootstrap 95\% interval of $[0.0212,0.0290]$~mm. It improves all 28 camera-pair means and cuts acceleration magnitude by 31.04\%.

\noindent\textbf{Body branch.}
With the body predictor and gate frozen, we instead fit four effective scalars of \Cref{eq:adaptive_temporal_variances}, namely $a_\nu=\lambda_\nu/\sigma_\nu^{2}$, $\sigma_{\mathrm{obs},0}^{2}$, $\lambda_{\mathrm{rel}}$, and $a_\Delta=\lambda_\Delta/\sigma_\Delta^{2}$, by minimizing reliable-13 PA-MPJPE over 105,126 consecutive training pairs from the gate-training subjects. Optimization uses Adam for 10 epochs with a learning rate of \(0.02\), and epoch~6 is selected by the same criterion on 4,662 S16 validation pairs. The fitted values are $a_\nu=31.4855$, $\sigma_{\mathrm{obs},0}^{2}=0.03987$, $\lambda_{\mathrm{rel}}=0.83239$, and $a_\Delta=3.56423$, with $\sigma_{\mathrm{proc},0}^{2}$ fixed to one.

On the S1, S9, and S11 test subjects, the fitted setup lowers GT15 PA-MPJPE from 76.9409~mm to 76.5428~mm, improving each subject. The resulting 0.3981~mm gain has a sequence-block-bootstrap 95\% interval of $[0.3012,0.5045]$~mm, and angular-acceleration magnitude falls from 1.8535 to 1.2569$^\circ$/frame$^2$ (32.19\%).

\subsection{Wearable-System Demonstration}
\label{sec:rig_evaluation}
According to the policy of the authors' institution, ethics review was not required for this study. This experiment tests whether the calibrated wearable produces uninterrupted trajectory and articulation outputs on a real recording.

\subsubsection{Outward-Looking Camera Tracking}
\label{sec:outward_tracking}

Following \Cref{sec:rig_trajectory}, the rig trajectory places the downward-pair estimates in a common world frame. The four outward streams---the $170^\circ$ forward pair (Cam1/2) and the $220^\circ$ lateral pair (Cam5/6)---are tracked as a single calibrated generalized camera, with the extrinsics of \Cref{sec:calibration} held fixed so that metric scale follows from the inter-camera baselines. Each view uses its calibrated Scaramuzza fisheye model~\cite{scaramuzza2006flexible}. ORB descriptors are aggregated across the four views for loop-closure detection.

\begin{figure}[tbp]
\centering
\includegraphics[width=\linewidth]{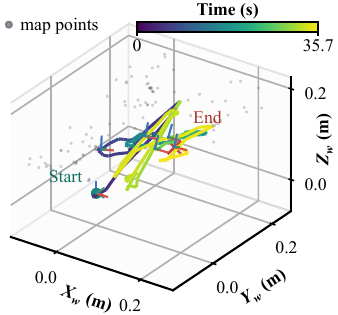}
\caption{World-frame rig trajectory and sparse map estimated from the four outward views, colored by time over the 35.7-s sequence. Sampled axes show rig orientation.}
\label{fig:outward_rig_trajectory}
\end{figure}

\noindent\textbf{Demonstration sequence.}
From a 35.7-s synchronized six-camera recording at $1920\times1080$ and 59.94~fps, the four outward streams track without loss and export 2,121 timestamp-aligned 6-DoF rig poses. Calibration reached $0.093$/$0.106$~px reprojection error. \Cref{fig:outward_rig_trajectory} shows the recovered trajectory and its sparse map, with the raw views in \Cref{fig:teaser}. During this recording, the wearer performs near-body interactions within a limited workspace. The estimated trajectory has a cumulative length of $4.635$~m, stays within $0.279$~m of the start, and ends $0.220$~m away. Inter-frame translation and rotation stay below $0.022$~m and $1.49^\circ$; over the sequence, orientation deviates from its initial value by at most $41.1^\circ$ as a geodesic angle on $SO(3)$.

\subsubsection{Real-World Body and Hand Motion Capture}

\begin{figure}[tbp]
\centering
\includegraphics[width=\linewidth]{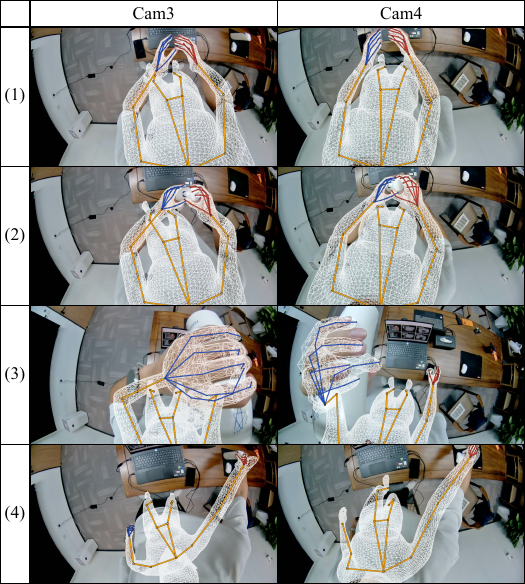}
\caption{Downward-pair output: no occlusion, hand--object interaction, hand--object occlusion, and one hand out of view. SMPL-X body (white) and MANO hands (blue left, red right) are reprojected into Cam3 and Cam4. The right hand in case (3) and the left hand in case (4) are seen in one view only, so they are single-view predictions.}
\label{fig:real_motion_capture}
\end{figure}

The same synchronized six-camera recording is processed end to end. Metric root placement uses the downward pair's $152.02$~mm baseline. Crops come from off-the-shelf detectors: the hand detector released with WiLoR~\cite{potamias2025wilor} gives the hand boxes and their left/right labels, and a Faster R-CNN~\cite{ren2015faster} gives the body box. Neither is fine-tuned, both boxes are enlarged by $1.2\times$ before cropping, and Cam3 and Cam4 are detected independently.

\noindent\textbf{Outputs.}
Calibrated rig transforms express the head-mounted rig, body, and both hands in one world frame (\Cref{fig:teaser}). \Cref{fig:real_motion_capture} projects the body and hand outputs back into the downward pair in four cases: (1) an unoccluded frame, (2) hand--object interaction, (3) hand--object occlusion, and (4) one hand outside the field of view. Cases (1) and (2) are the regime the paired-view fusion is designed for. In cases (3) and (4) one hand is missing from a single view, which is what the availability marking of \Cref{sec:data_flow} exists to handle: a hand missing from either view cannot be triangulated, so its metric placement is marked unavailable.

Detection is near-complete: Cam3 finds the left hand in 2,120 of the 2,121 frames and the right hand in 2,074, Cam4 in 2,069 and 2,048, and the body in every frame of both views. Triangulation therefore succeeds for the left hand in 2,069 frames and the right hand in 2,026, giving metric placement for both hands in 2,011 frames, for one hand in 73, and for neither in 37. For the MANO branch, the gate and temporal configuration are frozen from DexYCB validation, and no ground truth from this recording is used. Over valid consecutive paired-hand segments, per-joint soft fusion reaches an acceleration magnitude of $3.5866^\circ$/frame$^2$, which the adaptive $SO(3)$ refinement lowers to $2.1132^\circ$/frame$^2$, a reduction of 41.08\%. The gate stays active on this rig: Cam3 weights average $0.5116$ but span $0.098$ to $0.896$ across joints, with a mean adaptive gain of $0.7079$. This demonstration establishes synchronized output availability and end-to-end deployment.

\subsection{Limitations and Future Work}
\label{sec:limitations}

The benchmark protocols report similarity-aligned error, which solves for scale, rotation, and translation and therefore cannot evaluate absolute placement at all. They also lack our downward-pair geometry, and the wearable recording carries no external reference, so the articulation gains are measured under the standard protocols while the trajectory and placement stages are demonstrated rather than quantified. Most of the benchmark gain comes from using both views; the learned gates add a further increment over equal-weight fusion. On HO3Dv3 the frozen gate improves the mean and median but not P90 or P95. The MANO hands are placed at their independently triangulated roots rather than being kinematically reconciled with the SMPL-X arm chains, so forearm length is not enforced and the body and hand shape parameters may disagree. Root orientation and translation are not temporally filtered, and the pipeline is offline because temporal inconsistency uses a following frame and shape uses the full sequence.

Four directions follow from the present design. Reconciling the triangulated MANO root with the arm chain would make the exported skeleton kinematically consistent and let forearm length constrain the hand. Replacing the two-sided temporal cue and the sequence-level shape estimate with causal counterparts would make the pipeline online. A controlled capture with an external tracking reference would turn metric placement from a demonstrated capability into a measured one. Reducing weight toward eyeglasses scale, with temple-mounted outward cameras and overlapping oblique-downward views, would preserve the same role split.

\section{Conclusion}

We presented \method{}, a six-camera wearable and an offline pipeline that recovers body and bimanual hand motion by reliability-aware fusion and $SO(3)$ temporal refinement. Across UnrealEgo-RW, DexYCB, and HO3Dv3, dual-view fusion improves similarity-aligned local geometry, and temporal refinement improves all 28 DexYCB camera-pair means. The same recurrence applies to the body branch. On the rig itself, gates frozen from the benchmarks are used without retuning: metric placement is available for both hands in 2,011 of 2,121 frames, and refinement lowers the hand acceleration magnitude by 41.08\%. Together, these results support role-specialized sensing and canonical-space fusion.

\FloatBarrier
\bibliography{egowholeview_refs}

@inproceedings{prakash2024wildhands,
  title={3D Hand Pose Estimation in Everyday Egocentric Images},
  author={Prakash, Aditya and Tu, Ruisen and Chang, Matthew and Gupta, Saurabh},
  booktitle={European Conference on Computer Vision},
  year={2024}
}

@article{mucha2026vhpot,
  title={Towards Egocentric 3D Hand Pose Estimation in Unseen Domains},
  author={Mucha, Wiktor and Wray, Michael and Kampel, Martin},
  journal={arXiv preprint arXiv:2601.06537},
  year={2026}
}

@article{zhang2025hawor,
  title={HaWoR: World-Space Hand Motion Reconstruction from Egocentric Videos},
  author={Zhang, Jinglei and Deng, Jiankang and Ma, Chao and Potamias, Rolandos Alexandros},
  journal={arXiv preprint arXiv:2501.02973},
  year={2025}
}

@article{rim2025egoexo,
  title={Ego-Exo 3D Hand Tracking in the Wild with a Mobile Multi-Camera Rig},
  author={Rim, Patrick and He, Kun and Harris, Kevin and Copple, Braden and Han, Shangchen and An, Sizhe and Shugurov, Ivan and Hodan, Tomas and Wen, He and Xie, Xu},
  journal={arXiv preprint arXiv:2510.02601},
  year={2025}
}

@inproceedings{pavlakos2023hamer,
  title={Reconstructing Hands in 3D with Transformers},
  author={Pavlakos, Georgios and Shan, Dandan and Radosavovic, Ilija and Kanazawa, Angjoo and Fouhey, David and Malik, Jitendra},
  booktitle={Proceedings of the IEEE/CVF Conference on Computer Vision and Pattern Recognition},
  year={2024}
}

@article{potamias2025wilor,
  title={WiLoR: End-to-end 3D Hand Localization and Reconstruction in-the-wild},
  author={Potamias, Rolandos Alexandros and Zhang, Jinglei and Deng, Jiankang and Zafeiriou, Stefanos},
  journal={arXiv preprint arXiv:2409.12259},
  year={2025}
}

@inproceedings{ren2022mining,
  title={Mining Multi-View Information: A Strong Self-Supervised Framework for Depth-Based 3D Hand Pose and Mesh Estimation},
  author={Ren, Pengfei and Sun, Haifeng and Hao, Jiachang and Wang, Jingyu and Qi, Qi and Liao, Jianxin},
  booktitle={Proceedings of the IEEE/CVF Conference on Computer Vision and Pattern Recognition},
  year={2022}
}

@article{lin2024omnihands,
  title={OmniHands: Towards Robust 4D Hand Mesh Recovery via A Versatile Transformer},
  author={Lin, Dixuan and Zhang, Yuxiang and Li, Mengcheng and Liu, Yebin and Jing, Wei and Yan, Qi and Wang, Qianying and Zhang, Hongwen},
  journal={arXiv preprint arXiv:2405.20330},
  year={2024}
}

@article{zhang2025robowheel,
  title={RoboWheel: A Data Engine from Real-World Human Demonstrations for Cross-Embodiment Robotic Learning},
  author={Zhang, Yuhong and Gao, Zihan and Li, Shengpeng and Chen, Ling-Hao and Liu, Kaisheng and Cheng, Runqing and Lin, Xiao and Liu, Junjia and Li, Zhuoheng and Feng, Jingyi and He, Ziyan and Lin, Jintian and Huang, Zheyan and Liu, Zhifang and Wang, Haoqian},
  journal={arXiv preprint arXiv:2512.02729},
  year={2025}
}

@book{bowman2004dui,
  title={{3D} User Interfaces: Theory and Practice},
  author={Bowman, Doug A. and Kruijff, Ernst and LaViola, Joseph J. and Poupyrev, Ivan},
  publisher={Addison-Wesley},
  year={2004}
}

@inproceedings{grauman2022ego4d,
  title={{Ego4D}: Around the World in 3,000 Hours of Egocentric Video},
  author={Grauman, Kristen and Westbury, Andrew and Byrne, Eugene and Chavis, Zachary and Furnari, Antonino and Girdhar, Rohit and Hamburger, Jackson and Jiang, Hao and Liu, Miao and Liu, Xingyu and others},
  booktitle={Proceedings of the IEEE/CVF Conference on Computer Vision and Pattern Recognition},
  pages={18973--18990},
  year={2022}
}

@inproceedings{wang2023holoassist,
  title={{HoloAssist}: An Egocentric Human Interaction Dataset for Interactive {AI} Assistants in the Real World},
  author={Wang, Xin and Kwon, Taein and Rad, Mahdi and Pan, Bowen and Chakraborty, Ishani and Andrist, Sean and Bohus, Dan and Feniello, Ashley and Frujeri, Felipe Vieira and Joshi, Neel and Pollefeys, Marc},
  booktitle={Proceedings of the IEEE/CVF International Conference on Computer Vision},
  year={2023}
}

@inproceedings{zhang2022egobody,
  title={{EgoBody}: Human Body Shape and Motion of Interacting People from Head-Mounted Devices},
  author={Zhang, Siwei and Ma, Qianli and Zhang, Yan and Qian, Zhiyin and Kwon, Taein and Pollefeys, Marc and Bogo, Federica and Tang, Siyu},
  booktitle={Proceedings of the European Conference on Computer Vision},
  year={2022}
}

@article{han2020megatrack,
  title={{MEgATrack}: Monochrome Egocentric Articulated Hand-Tracking for Virtual Reality},
  author={Han, Shangchen and Liu, Beibei and Cabezas, Randi and Twigg, Christopher D. and Zhang, Peizhao and Petkau, Jeff and Yu, Tsz-Ho and Tai, Chun-Jung and Akbay, Muzaffer and Wang, Zheng and Nitzan, Asaf and Dong, Gang and Ye, Yuting and Tao, Lingling and Wan, Chengde and Wang, Robert},
  journal={ACM Transactions on Graphics},
  volume={39},
  number={4},
  year={2020}
}

@article{han2022umetrack,
  title={{UmeTrack}: Unified Multi-View End-to-End Hand Tracking for {VR}},
  author={Han, Shangchen and Wu, Po-Chen and Zhang, Yubo and Liu, Beibei and Zhang, Linguang and Wang, Zheng and Si, Weiguang and Zhang, Peizhao and Cai, Yujun and Hodan, Tomas and Cabezas, Randi and Tran, Luan and Akbay, Muzaffer and Yu, Tsz-Ho and Keskin, Cem and Wang, Robert},
  journal={ACM Transactions on Graphics},
  volume={41},
  number={6},
  year={2022}
}

@article{xu2019mo2cap2,
  title={{Mo2Cap2}: Real-Time Mobile {3D} Motion Capture with a Cap-Mounted Fisheye Camera},
  author={Xu, Weipeng and Chatterjee, Avishek and Zollhoefer, Michael and Rhodin, Helge and Fua, Pascal and Seidel, Hans-Peter and Theobalt, Christian},
  journal={IEEE Transactions on Visualization and Computer Graphics},
  volume={25},
  number={5},
  pages={2093--2101},
  year={2019}
}

@inproceedings{tome2019xregopose,
  title={{xR-EgoPose}: Egocentric {3D} Human Pose from an {HMD} Camera},
  author={Tome, Denis and Peluse, Patrick and Agapito, Lourdes and Badino, Hernan},
  booktitle={Proceedings of the IEEE/CVF International Conference on Computer Vision},
  year={2019}
}

@inproceedings{zhao2021egoglass,
  title={{EgoGlass}: Egocentric-View Human Pose Estimation from an Eyeglass Frame},
  author={Zhao, Dongxu and Wei, Zhen and Mahmud, Jisan and Frahm, Jan-Michael},
  booktitle={Proceedings of the International Conference on 3D Vision},
  pages={32--41},
  year={2021}
}

@inproceedings{akada2022unrealego,
  title={{UnrealEgo}: A New Dataset for Robust Egocentric {3D} Human Motion Capture},
  author={Akada, Hiroyasu and Wang, Jian and Shimada, Soshi and Takahashi, Masaki and Theobalt, Christian and Golyanik, Vladislav},
  booktitle={Proceedings of the European Conference on Computer Vision},
  year={2022}
}

@article{romero2017embodied,
  title={Embodied Hands: Modeling and Capturing Hands and Bodies Together},
  author={Romero, Javier and Tzionas, Dimitrios and Black, Michael J.},
  journal={ACM Transactions on Graphics},
  volume={36},
  number={6},
  year={2017}
}

@inproceedings{pavlakos2019expressive,
  title={Expressive Body Capture: {3D} Hands, Face, and Body from a Single Image},
  author={Pavlakos, Georgios and Choutas, Vasileios and Ghorbani, Nima and Bolkart, Timo and Osman, Ahmed A. A. and Tzionas, Dimitrios and Black, Michael J.},
  booktitle={Proceedings of the IEEE/CVF Conference on Computer Vision and Pattern Recognition},
  year={2019}
}

@article{ionescu2014human36m,
  title={{Human3.6M}: Large Scale Datasets and Predictive Methods for {3D} Human Sensing in Natural Environments},
  author={Ionescu, Catalin and Papava, Dragos and Olaru, Vlad and Sminchisescu, Cristian},
  journal={IEEE Transactions on Pattern Analysis and Machine Intelligence},
  volume={36},
  number={7},
  pages={1325--1339},
  year={2014}
}

@inproceedings{kocabas2020vibe,
  title={{VIBE}: Video Inference for Human Body Pose and Shape Estimation},
  author={Kocabas, Muhammed and Athanasiou, Nikos and Black, Michael J.},
  booktitle={Proceedings of the IEEE/CVF Conference on Computer Vision and Pattern Recognition},
  year={2020}
}

@inproceedings{hampali2020honnotate,
  title={{HOnnotate}: A Method for {3D} Annotation of Hand and Object Poses},
  author={Hampali, Shreyas and Rad, Mahdi and Oberweger, Markus and Lepetit, Vincent},
  booktitle={Proceedings of the IEEE/CVF Conference on Computer Vision and Pattern Recognition},
  year={2020}
}

@inproceedings{chao2021dexycb,
  title={{DexYCB}: A Benchmark for Capturing Hand Grasping of Objects},
  author={Chao, Yu-Wei and Yang, Wei and Xiang, Yu and Molchanov, Pavlo and Handa, Ankur and Tremblay, Jonathan and Narang, Yashraj S. and Van Wyk, Karl and Iqbal, Umar and Birchfield, Stan and Kautz, Jan and Fox, Dieter},
  booktitle={Proceedings of the IEEE/CVF Conference on Computer Vision and Pattern Recognition},
  year={2021}
}

@inproceedings{banerjee2024hot3d,
  title={{HOT3D}: Hand and Object Tracking in {3D} from Egocentric Multi-View Videos},
  author={Banerjee, Prithviraj and Shkodrani, Sindi and Moulon, Pierre and Hampali, Shreyas and Han, Shangchen and Zhang, Fan and Zhang, Linguang and Fountain, Jade and Miller, Edward and Basol, Selen and Newcombe, Richard and Wang, Robert and Engel, Jakob Julian and Hodan, Tomas},
  booktitle={Proceedings of the IEEE/CVF Conference on Computer Vision and Pattern Recognition},
  pages={7061--7071},
  year={2025}
}

@inproceedings{nair2022r3m,
  title={{R3M}: A Universal Visual Representation for Robot Manipulation},
  author={Nair, Suraj and Rajeswaran, Aravind and Kumar, Vikash and Finn, Chelsea and Gupta, Abhinav},
  booktitle={Proceedings of the Conference on Robot Learning},
  year={2022}
}

@article{brohan2023rt2,
  title={{RT-2}: Vision-Language-Action Models Transfer Web Knowledge to Robotic Control},
  author={Brohan, Anthony and Brown, Noah and Carbajal, Justice and Chebotar, Yevgen and Chen, Xi and Choromanski, Krzysztof and Ding, Tianli and Driess, Danny and Dubey, Avinava and Finn, Chelsea and others},
  journal={arXiv preprint arXiv:2307.15818},
  year={2023}
}

@inproceedings{khazatsky2024droid,
  title={{DROID}: A Large-Scale In-the-Wild Robot Manipulation Dataset},
  author={Khazatsky, Alexander and Pertsch, Karl and Nair, Suraj and Balakrishna, Ashwin and Dasari, Sudeep and Karamcheti, Siddharth and Nasiriany, Soroush and Srirama, Mohan Kumar and Chen, Lawrence Yunliang and Ellis, Kirsty and others},
  booktitle={Proceedings of Robotics: Science and Systems},
  year={2024}
}

@article{xu2026surfacexr,
  title={{SurfaceXR}: Fusing Smartwatch {IMUs} and Egocentric Hand Pose for Seamless Surface Interactions},
  author={Xu, Vasco and Chen, Brian and Gonzalez, Eric J. and Cola{\c{c}}o, Andrea and Hoffmann, Henry and Gonzalez-Franco, Mar and Ahuja, Karan},
  journal={IEEE Transactions on Visualization and Computer Graphics},
  year={2026},
  note={IEEE VR journal track; arXiv:2603.19529}
}

@inproceedings{cho2026unifiedho,
  title={A Unified Hand and Gesture Tracking via Offloading Framework for Object-mediated Interaction in Wearable {AR}},
  author={Cho, Woojin and Ha, Taewook and Son, Taejun and Woo, Woontack},
  booktitle={Proceedings of the IEEE Conference on Virtual Reality and 3D User Interfaces},
  year={2026}
}

@inproceedings{liu2022hoi4d,
  title={{HOI4D}: A 4D Egocentric Dataset for Category-Level Human-Object Interaction},
  author={Liu, Yunze and Liu, Yun and Jiang, Che and Lyu, Kangbo and Wan, Weikang and Shen, Hao and Liang, Boqiang and Fu, Zhoujie and Wang, He and Yi, Li},
  booktitle={Proceedings of the IEEE/CVF Conference on Computer Vision and Pattern Recognition},
  year={2022}
}

@article{na2026egowholebody,
  title={Egocentric Whole-Body Human Mesh Recovery with Prior-Guided Learning},
  author={Na, Soyeon and Noh, Seung Young and Chang, Ju Yong},
  journal={arXiv preprint arXiv:2605.08606},
  year={2026}
}

@article{wang2024cylindertag,
  author    = {Shaoan Wang and Mingzhu Zhu and Yaoqing Hu and Dongyue Li and
               Fusong Yuan and Junzhi Yu},
  title     = {{CylinderTag}: An Accurate and Flexible Marker for Cylinder-Shape
               Objects Pose Estimation Based on Projective Invariants},
  journal   = {IEEE Transactions on Visualization and Computer Graphics},
  year      = {2024},
  month     = dec,
  volume    = {30},
  number    = {12},
  pages     = {7486--7499},
  doi       = {10.1109/TVCG.2024.3350901},
  url       = {https://ieeexplore.ieee.org/document/10382637},
  publisher = {IEEE}
}

@article{hu2024lightweightnavigation,
  author    = {Yaoqing Hu and Mingzhu Zhu and Shaoan Wang and Dongyue Li and
               Fusong Yuan and Junzhi Yu},
  title     = {A Novel Lightweight Navigation System for Oral and Maxillofacial
               Surgery Using an External Curved Self-Identifying Checkerboard},
  journal   = {IEEE Transactions on Automation Science and Engineering},
  year      = {2024},
  month     = apr,
  volume    = {21},
  number    = {2},
  pages     = {1434--1444},
  doi       = {10.1109/TASE.2023.3241325},
  url       = {https://ieeexplore.ieee.org/document/10038496},
  publisher = {IEEE}
}

@article{campos2021orbslam3,
  author  = {Campos, Carlos and Elvira, Richard and G{\'o}mez Rodr{\'i}guez, Juan J. and Montiel, Jos{\'e} M. M. and Tard{\'o}s, Juan D.},
  title   = {{ORB-SLAM3}: An Accurate Open-Source Library for Visual, Visual-Inertial and Multi-Map {SLAM}},
  journal = {IEEE Transactions on Robotics},
  volume  = {37},
  number  = {6},
  pages   = {1874--1890},
  year    = {2021},
  doi     = {10.1109/TRO.2021.3075644}
}

@article{urban2016multicolslam,
  author  = {Urban, Steffen and Hinz, Stefan},
  title   = {{MultiCol-SLAM}: A Modular Real-Time Multi-Camera {SLAM} System},
  journal = {arXiv preprint arXiv:1610.07336},
  year    = {2016},
  url     = {https://arxiv.org/abs/1610.07336}
}

@article{demayo2025monado,
  author  = {de Mayo, Mateo and Cremers, Daniel and Pire, Taih{\'u}},
  title   = {The {Monado} {SLAM} Dataset for Egocentric Visual-Inertial Tracking},
  journal = {arXiv preprint arXiv:2508.00088},
  year    = {2025},
  url     = {https://arxiv.org/abs/2508.00088}
}

@misc{ropedia2026xperience10m,
  author       = {{Ropedia}},
  title        = {{Xperience-10M}: A Large-Scale Egocentric Multimodal Dataset with Structured {3D/4D} Annotations},
  howpublished = {Hugging Face dataset},
  month        = mar,
  year         = {2026},
  url          = {https://huggingface.co/datasets/ropedia-ai/xperience-10m},
  note         = {Accessed July 12, 2026}
}

@article{lin2026geohand,
  title={{GeoHand}: Unlocking Prior Geometry Knowledge for Monocular 3D Hand Reconstruction},
  author={Lin, Weiquan and Hu, Yaoqing and Dai, Liangchen and Tang, Xu and Chen, Xingyu},
  journal={arXiv preprint arXiv:2605.17354},
  year={2026}
}

@inproceedings{kolotouros2021prohmr,
  title={Probabilistic Modeling for Human Mesh Recovery},
  author={Kolotouros, Nikos and Pavlakos, Georgios and Jayaraman, Dinesh and Daniilidis, Kostas},
  booktitle={Proceedings of the IEEE/CVF International Conference on Computer Vision},
  pages={11605--11614},
  year={2021}
}

@inproceedings{sengupta2023humaniflow,
  title={{HuManiFlow}: Ancestor-Conditioned Normalising Flows on {$SO(3)$} Manifolds for Human Pose and Shape Distribution Estimation},
  author={Sengupta, Akash and Budvytis, Ignas and Cipolla, Roberto},
  booktitle={Proceedings of the IEEE/CVF Conference on Computer Vision and Pattern Recognition},
  pages={4779--4789},
  year={2023}
}

@inproceedings{dwivedi2024poco,
  title={{POCO}: 3D Pose and Shape Estimation with Confidence},
  author={Dwivedi, Sai Kumar and Schmid, Cordelia and Yi, Hongwei and Black, Michael J. and Tzionas, Dimitrios},
  booktitle={International Conference on 3D Vision},
  pages={85--95},
  year={2024},
  doi={10.1109/3DV62453.2024.00115}
}

@inproceedings{bramlage2023plausible,
  title={Plausible Uncertainties for Human Pose Regression},
  author={Bramlage, Lennart and Karg, Michelle and Curio, Cristobal},
  booktitle={Proceedings of the IEEE/CVF International Conference on Computer Vision},
  pages={15133--15142},
  year={2023}
}

@inproceedings{kocabas2021pare,
  title={{PARE}: Part Attention Regressor for 3D Human Body Estimation},
  author={Kocabas, Muhammed and Huang, Chun-Hao P. and Hilliges, Otmar and Black, Michael J.},
  booktitle={Proceedings of the IEEE/CVF International Conference on Computer Vision},
  pages={11127--11137},
  year={2021}
}

@inproceedings{park2022handoccnet,
  title={{HandOccNet}: Occlusion-Robust 3D Hand Mesh Estimation Network},
  author={Park, JoonKyu and Oh, Yeonguk and Moon, Gyeongsik and Choi, Hongsuk and Lee, Kyoung Mu},
  booktitle={Proceedings of the IEEE/CVF Conference on Computer Vision and Pattern Recognition},
  pages={1496--1505},
  year={2022}
}

@inproceedings{iskakov2019learnable,
  title={Learnable Triangulation of Human Pose},
  author={Iskakov, Karim and Burkov, Egor and Lempitsky, Victor and Malkov, Yury},
  booktitle={Proceedings of the IEEE/CVF International Conference on Computer Vision},
  pages={7718--7727},
  year={2019}
}

@article{engel2023projectaria,
  title={{Project Aria}: A New Tool for Egocentric Multi-Modal {AI} Research},
  author={Engel, Jakob and Somasundaram, Kiran and Goesele, Michael and others},
  journal={arXiv preprint arXiv:2308.13561},
  year={2023}
}

@inproceedings{ma2024nymeria,
  title={{Nymeria}: A Massive Collection of Multimodal Egocentric Daily Motion in the Wild},
  author={Ma, Lingni and Ye, Yuting and Hong, Fangzhou and Guzov, Vladimir and Jiang, Yifeng and Postyeni, Rowan and Pesqueira, Luis and Gamino, Alexander and Baiyya, Vijay and Kim, Hyo Jin and Bailey, Kevin and Fosas, David Soriano and Liu, C. Karen and Liu, Ziwei and Engel, Jakob and De Nardi, Renzo and Newcombe, Richard},
  booktitle={Proceedings of the European Conference on Computer Vision},
  pages={445--465},
  year={2024},
  doi={10.1007/978-3-031-72691-0_25}
}

@inproceedings{yi2025egoallo,
  title={Estimating Body and Hand Motion in an Ego-sensed World},
  author={Yi, Brent and Ye, Vickie and Zheng, Maya and Li, Yunqi and M{\"u}ller, Lea and Pavlakos, Georgios and Ma, Yi and Malik, Jitendra and Kanazawa, Angjoo},
  booktitle={Proceedings of the IEEE/CVF Conference on Computer Vision and Pattern Recognition},
  pages={7072--7084},
  year={2025}
}

@inproceedings{liu2024s2dhand,
  title={Single-to-Dual-View Adaptation for Egocentric {3D} Hand Pose Estimation},
  author={Liu, Ruicong and Ohkawa, Takehiko and Zhang, Mingfang and Sato, Yoichi},
  booktitle={Proceedings of the IEEE/CVF Conference on Computer Vision and Pattern Recognition},
  pages={677--686},
  year={2024}
}

@inproceedings{tse2023spectral,
  title={Spectral Graphormer: Spectral Graph-Based Transformer for Egocentric Two-Hand Reconstruction using Multi-View Color Images},
  author={Tse, Tze Ho Elden and Mueller, Franziska and Shen, Zhengyang and Tang, Danhang and Beeler, Thabo and Dou, Mingsong and Zhang, Yinda and Petrovic, Sasa and Chang, Hyung Jin and Taylor, Jonathan and Doosti, Bardia},
  booktitle={Proceedings of the IEEE/CVF International Conference on Computer Vision},
  pages={14666--14677},
  year={2023}
}

@inproceedings{yang2023poem,
  title={{POEM}: Reconstructing Hand in a Point Embedded Multi-View Stereo},
  author={Yang, Lixin and Xu, Jian and Zhong, Licheng and Zhan, Xinyu and Wang, Zhicheng and Wu, Kejian and Lu, Cewu},
  booktitle={Proceedings of the IEEE/CVF Conference on Computer Vision and Pattern Recognition},
  pages={21108--21117},
  year={2023}
}

@inproceedings{zheng2023hamuco,
  title={{HaMuCo}: Hand Pose Estimation via Multiview Collaborative Self-Supervised Learning},
  author={Zheng, Xiaozheng and Wen, Chao and Xue, Zhou and Ren, Pengfei and Wang, Jingyu},
  booktitle={Proceedings of the IEEE/CVF International Conference on Computer Vision},
  pages={20763--20773},
  year={2023}
}

@inproceedings{ren2025rule,
  title={Rule Meets Learning: Confidence-Aware Multi-View Fusion for Self-Supervised {3D} Hand Pose Estimation},
  author={Ren, Pengfei and Wang, Jingyu and Sun, Haifeng and Qi, Qi and Wang, Jing and Liao, Jianxin},
  booktitle={Proceedings of the 33rd ACM International Conference on Multimedia},
  pages={1646--1655},
  year={2025},
  doi={10.1145/3746027.3755342}
}

@inproceedings{wang2022egopw,
  title={Estimating Egocentric 3D Human Pose in the Wild With External Weak Supervision},
  author={Wang, Jian and Liu, Lingjie and Xu, Weipeng and Sarkar, Kripasindhu and Luvizon, Diogo and Theobalt, Christian},
  booktitle={Proceedings of the IEEE/CVF Conference on Computer Vision and Pattern Recognition},
  pages={13157--13166},
  year={2022}
}

@inproceedings{wang2023sceneego,
  title={Scene-aware Egocentric 3D Human Pose Estimation},
  author={Wang, Jian and Luvizon, Diogo and Xu, Weipeng and Liu, Lingjie and Sarkar, Kripasindhu and Theobalt, Christian},
  booktitle={Proceedings of the IEEE/CVF Conference on Computer Vision and Pattern Recognition},
  year={2023}
}

@article{kilteni2012embodiment,
  title={The Sense of Embodiment in Virtual Reality},
  author={Kilteni, Konstantina and Groten, Raphaela and Slater, Mel},
  journal={Presence: Teleoperators and Virtual Environments},
  volume={21},
  number={4},
  pages={373--387},
  year={2012}
}

@article{slater2010bodytransfer,
  title={First Person Experience of Body Transfer in Virtual Reality},
  author={Slater, Mel and Spanlang, Bernhard and Sanchez-Vives, Maria V. and Blanke, Olaf},
  journal={PLoS ONE},
  volume={5},
  number={5},
  pages={e10564},
  year={2010},
  doi={10.1371/journal.pone.0010564}
}

@article{waltemate2018avatar,
  title={The Impact of Avatar Personalization and Immersion on Virtual Body Ownership, Presence, and Emotional Response},
  author={Waltemate, Thomas and Gall, Dominik and Roth, Daniel and Botsch, Mario and Latoschik, Marc Erich},
  journal={IEEE Transactions on Visualization and Computer Graphics},
  volume={24},
  number={4},
  pages={1643--1652},
  year={2018},
  doi={10.1109/TVCG.2018.2794629}
}

@inproceedings{zhang2018vrteleop,
  title={Deep Imitation Learning for Complex Manipulation Tasks from Virtual Reality Teleoperation},
  author={Zhang, Tianhao and McCarthy, Zoe and Jow, Owen and Lee, Dennis and Chen, Xi and Goldberg, Ken and Abbeel, Pieter},
  booktitle={Proceedings of the IEEE International Conference on Robotics and Automation},
  year={2018}
}

@inproceedings{eubanks2020trackingfidelity,
  title={The Effects of Body Tracking Fidelity on Embodiment of an Inverse-Kinematic Avatar for Male Participants},
  author={Eubanks, Jessie Colette and Moore, Alec G. and Fishwick, Paul A. and McMahan, Ryan P.},
  booktitle={Proceedings of the IEEE International Symposium on Mixed and Augmented Reality},
  pages={54--63},
  year={2020},
  doi={10.1109/ISMAR50242.2020.00025}
}

@inproceedings{kim2024selfavatar,
  title={Self-Avatar Recognition in Virtual Reality: the Self Advantage Phenomenon and the Relative Importance of Motion and Visual Congruence},
  author={Kim, Bowon and Hwang, Eugene and Lee, Jeongmi},
  booktitle={Proceedings of the IEEE International Symposium on Mixed and Augmented Reality},
  pages={643--651},
  year={2024},
  doi={10.1109/ISMAR62088.2024.00079}
}

@article{maiorca2024artifacts,
  title={Self-Avatar Animation in Virtual Reality: Impact of Motion Signals Artifacts on the Full-Body Pose Reconstruction},
  author={Maiorca, Antoine and Ghasemzadeh, Seyed Abolfazl and Ravet, Thierry and Cresson, Fran{\c{c}}ois and Dutoit, Thierry and De Vleeschouwer, Christophe},
  journal={arXiv preprint arXiv:2404.18628},
  year={2024},
  doi={10.48550/arXiv.2404.18628}
}

@inproceedings{scaramuzza2006flexible,
  title={A Flexible Technique for Accurate Omnidirectional Camera Calibration and Structure from Motion},
  author={Scaramuzza, Davide and Martinelli, Agostino and Siegwart, Roland},
  booktitle={Proceedings of the IEEE International Conference on Computer Vision Systems},
  pages={45},
  year={2006},
  doi={10.1109/ICVS.2006.3}
}

@article{kalman1960new,
  title={A New Approach to Linear Filtering and Prediction Problems},
  author={Kalman, Rudolf E.},
  journal={Journal of Basic Engineering},
  volume={82},
  number={1},
  pages={35--45},
  year={1960},
  doi={10.1115/1.3662552}
}

@inproceedings{casiez2012oneeuro,
  title={1 Euro Filter: A Simple Speed-Based Low-Pass Filter for Noisy Input in Interactive Systems},
  author={Casiez, G{\'e}ry and Roussel, Nicolas and Vogel, Daniel},
  booktitle={Proceedings of the SIGCHI Conference on Human Factors in Computing Systems},
  pages={2527--2530},
  year={2012},
  doi={10.1145/2207676.2208639}
}

@article{park1997smooth,
  title={Smooth Invariant Interpolation of Rotations},
  author={Park, Frank C. and Ravani, Bahram},
  journal={ACM Transactions on Graphics},
  volume={16},
  number={3},
  pages={277--295},
  year={1997},
  doi={10.1145/256157.256160}
}

@inproceedings{ren2015faster,
  title     = {Faster {R-CNN}: Towards Real-Time Object Detection with Region Proposal Networks},
  author    = {Ren, Shaoqing and He, Kaiming and Girshick, Ross and Sun, Jian},
  booktitle = {Advances in Neural Information Processing Systems},
  pages     = {91--99},
  year      = {2015}
}
\end{document}